\documentclass[12pt, a4paper,fleqn]{cas-sc}
\newcommand{\printorcid}{}
\usepackage{algorithm}
\usepackage{xcolor}
\usepackage{algpseudocode}
\usepackage{graphicx}
\usepackage{subcaption}
\usepackage{float}
\usepackage{enumitem}
\usepackage{setspace}
\usepackage{adjustbox}
\usepackage{multirow}
\usepackage{wasysym}
\usepackage{pifont}
\usepackage[utf8]{inputenc}
\usepackage{amsmath}
\usepackage{amsfonts}
\usepackage{booktabs}
\usepackage{lipsum}
\definecolor{MineShaft}{rgb}{0.2,0.2,0.2}
\definecolor{darkgreen}{rgb}{0.0, 0.7, 0.0}
\def\tsc#1{\csdef{#1}{\textsc{\lowercase{#1}}\xspace}}
\tsc{WGM}
\tsc{QE}
\begin{document}
\let\WriteBookmarks\relax
\def\floatpagepagefraction{1}
\def\textpagefraction{.001}

\shorttitle{}    

\shortauthors{Yu et al.}  

\title [mode = title]{Dual-Splitting Conformal Prediction for Multi-Step Time Series Forecasting}  

% Authors
\author[1]{Qingdi Yu}
\author[2]{Zhiwei Cao}
\author[2]{Ruihang Wang}
\author[1]{Zhen Yang}
\author[1]{Lijun Deng}
\author[1]{Min Hu}
\author[3]{Yong Luo}
\author[1]{Xin Zhou}
\cormark[1]
\ead{zhouxin@jxstnu.edu.cn}

\affiliation[1]{organization={Jiangxi Science and Technology Normal University},
            addressline={589 Xuefu Avenue, Hongjiaozhou},
            city={Nanchang},
            postcode={330038}, 
            state={Jiangxi},
            country={China}}
            
\affiliation[2]{organization={Nanyang Technological University},
            addressline={50 Nanyang Avenue}, 
            city={Singapore},
            postcode={639798}, 
            country={Singapore}}
            
\affiliation[3]{organization={Wuhan University},
            addressline={No.299, Bayi Road, Wuchang District}, 
            city={Wuhan},
            postcode={430072}, 
            state={Hubei},
            country={China}}

\cortext[0]{The first two authors contribute equally.}
\cortext[1]{Corresponding author.}


\begin{abstract}
Time series forecasting is crucial for applications like resource scheduling and risk management, where multi-step predictions provide a comprehensive view of future trends. Uncertainty Quantification (UQ) is a mainstream approach for addressing forecasting uncertainties, with Conformal Prediction (CP) gaining attention due to its model-agnostic nature and statistical guarantees. However, most variants of CP are designed for single-step predictions and face challenges in multi-step scenarios, such as reliance on real-time data and limited scalability. This highlights the need for CP methods specifically tailored to multi-step forecasting.
We propose the Dual-Splitting Conformal Prediction (DSCP) method, a novel CP approach designed to capture inherent dependencies within time-series data for \textbf{multi-step} forecasting.
Experimental results on real-world datasets from four different domains demonstrate that the proposed DSCP significantly outperforms existing CP variants in terms of the Winkler Score, achieving a performance improvement of up to 23.59\% compared to state-of-the-art methods. Furthermore, we deployed the DSCP approach for renewable energy generation and IT load forecasting in power management of a real-world trajectory-based application, achieving an 11.25\% reduction in carbon emissions through predictive optimization of data center operations and controls.
\end{abstract}

% Keywords

\maketitle
% Main text
\section{Introduction}
Time series forecasting plays a crucial role in numerous real-world applications, ranging from resource scheduling and risk management to strategic planning. In these domains, accurate predictions are essential for making informed decisions. Time series forecasting methods are broadly categorized into single-step and multi-step predictions. Many real-world scenarios rely on long-term predictions, leading to an increase in the use of multi-step forecasting~\cite{liu2024short,dolgintseva2024comparison}. A growing body of literature has demonstrated that multi-step predictions can provide valuable insights into future trends, particularly in complicated cases like resource scheduling~\cite{ghobadi2022multi,yang2021comprehensive,priya2019resource}, risk management~\cite{he2024data,zhang2021multi}, and strategy creation~\cite{lim2021time,zhou2018predicting}. Therefore, advancements in time series forecasting techniques that focus on multi-step predictions align better with practical application requirements.

In this context, Uncertainty Quantification (UQ)~\cite{abdar2021review} has emerged as a critical tool for addressing the inherent uncertainties in time series forecasting. In recent years, UQ has been widely applied in various time series forecasting-related domains, including system control optimization~\cite{kabir2018neural}. By quantifying potential risks, UQ provides reliable information that facilitates the formulation of robust control policies in complex systems. Moreover, it enhances data presentation and result interpretation, offering decision-makers deeper insights into model reliability. As a consequence, UQ has established itself as a vital component in optimization and decision-making processes. Numerous studies have demonstrated the effectiveness of UQ in fields such as load forecasting~\cite{almeida2015prediction, quan2014uncertainty, quan2013short}, financial applications~\cite{tseng2010comparing, zhang2007statistical}, renewable energy forecasting~\cite{pinson2010conditional, galvan2017multi,wang2023conformal}, and medical applications~\cite{nishiura2012early, zee2016stroke, hernandez2020uncertainty}.

Among UQ methods, we choose Conformal Prediction (CP)~\cite{shafer2008tutorial} as the approach for quantifying uncertainty in multi-step time series forecasting. CP stands out due to its unique advantages: it provides statistically guaranteed coverage, is model-agnostic, and is non-parametric~\cite{fontana2023conformal}. Compared to Bayesian methods~\cite{box2011bayesian}, Monte Carlo methods, and deep learning approaches, CP offers higher computational efficiency. Additionally, CP is more interpretable than deep learning methods, fuzzy theory, and Gaussian process regression. Furthermore, CP adapts better to complex data patterns and model structures compared to interval analysis, fuzzy theory, and Bayesian methods. These characteristics make CP a transparent, practical, and robust choice for uncertainty quantification in multi-step time series forecasting.

Since CP and most CP variants are primarily designed for uncertainty quantification in single-step time series forecasting, modifications are needed to adapt them to multi-step forecasting scenarios. Despite being initially developed for non-time-series forecasting, CP has gained increasing attention in time series forecasting in recent years due to its flexibility and statistical guarantees. To address the challenges posed by sequential dependencies, several improved variants of CP have been proposed, including methods for updating error sets~\cite{xu2021conformal, barber2023conformal}, modifying hyperparameters~\cite{gibbs2021adaptive, sesia2021conformal, bastani2022practical, lin2022conformal, angelopoulos2024conformal}, customizing predictive models~\cite{koenker1978regression, xu2023sequential}, and categorizing data~\cite{stankeviciute2021conformal, sun2022copula, auer2024conformal}. However, these improved methods are mainly applicable to single-step prediction uncertainty quantification and cannot be effectively applied to real-world scenarios requiring multi-step prediction uncertainty measurement. Specifically, existing methods face two main challenges: 
\begin{itemize}[leftmargin=*, itemsep=0pt, parsep=1pt]
    \item They usually rely on real data acquired in real time to measure single-step prediction uncertainty, which prevents them from effectively measuring multi-step prediction uncertainty through recursive calls.
    \item These methods are designed to focus primarily on single-step prediction outputs, and their architectures are not well-suited for multi-step prediction tasks.
\end{itemize} 
As a result, CP variants are difficult to be used in multi-step prediction to meet practical needs.

In this paper, we propose Dual-Splitting Conformal Prediction (DSCP), a dedicated approach for multi-step time-series forecasting. The core idea of DSCP is to separately treat error information obtained from different conditions. This separation prevents interference between errors originating from distinct distributions, thereby improving the accuracy and reliability of uncertainty quantification in multi-step predictions. We compare it with improved variants of CP adapted for multi-step forecasting to enable a fair comparison with DSCP. Our contributions can be summarized in the following three points:
\begin{enumerate}[leftmargin=*, itemsep=0pt, parsep=1pt]
\item We propose the DSCP approach for measuring uncertainty in multi-step time-series forecasting. The core innovation of DSCP lies in its dual-dimensional split of the error set, which separates error information with different distributions, enabling more accurate uncertainty quantification.
\item We made appropriate modifications to the improved variants of CP to enable a performance comparison of uncertainty measurement with DSCP on multi-step time series data. These modifications were made while preserving the core operations of those methods. The results show that DSCP outperforms the improved variants of CP, achieving an average performance improvement of 11.08\% and a maximum improvement of 23.59\%.
\item We apply DSCP to a real-world trajectory based application in system optimization. Experimental results demonstrate that, in a simulated data center application environment, the DSCP method achieved an average 8.05\% reduction in carbon emissions compared with the baseline, with the best-case scenario showing a 11.25\% reduction in carbon emissions.
\end{enumerate}

The remainder of this paper is organized as follows. Section 2 reviews related work on improved variants of CP for time series forecasting. Section 3 introduces the DSCP approach and its core methodology. Section 4 presents the experimental setup and comparative results. Section 5 demonstrates DSCP’s practical application in data center energy management. Section 6 concludes the paper and discusses future work. 

\section{Related Work}
This section provides an overview of the improved CP variants used in time series forecasting. We categorize the improved CP variants into four main approaches, with a detailed explanation of the core improvement strategies underlying each category. Additionally, the final part of this chapter summarizes the limitations of these methods.

\subsection{Overview of CP}
CP is a powerful UQ framework that provides reliable prediction intervals for point predictions. Point predictions, generated by models such as regression, machine learning, or neural networks, yield single-valued forecasts at each time step. CP is model-agnostic, meaning it can be applied to any predictive model regardless of its internal structure.

CP constructs prediction intervals that guarantee a specified probability of containing the true values. This makes CP particularly useful for uncertainty quantification in complex models like deep learning, where traditional methods often rely on restrictive assumptions about data distribution. CP offers a more flexible and transparent approach by leveraging error sets derived from model predictions.

\subsection{Application of CP in Time Series}
While CP has been successfully applied in domains like classification and regression, its application to time-series forecasting introduces additional complexities. Time-series data exhibit temporal dependencies that make uncertainty quantification more challenging. Specifically, when working with sequential data, the dependencies between consecutive observations need to be carefully considered to ensure the prediction intervals remain accurate and reliable.

Despite these challenges, recent advancements have led to the application of CP to time-series forecasting. Researchers have recognized the potential of CP for time-series data and have developed approaches to adapt its workflow to handle the inherent complexities of sequential predictions. These adaptations aim to address temporal dependencies and enhance the accuracy of uncertainty quantification in time-series forecasting.

\begin{figure}
    \centering
    \includegraphics[width=0.8\linewidth]{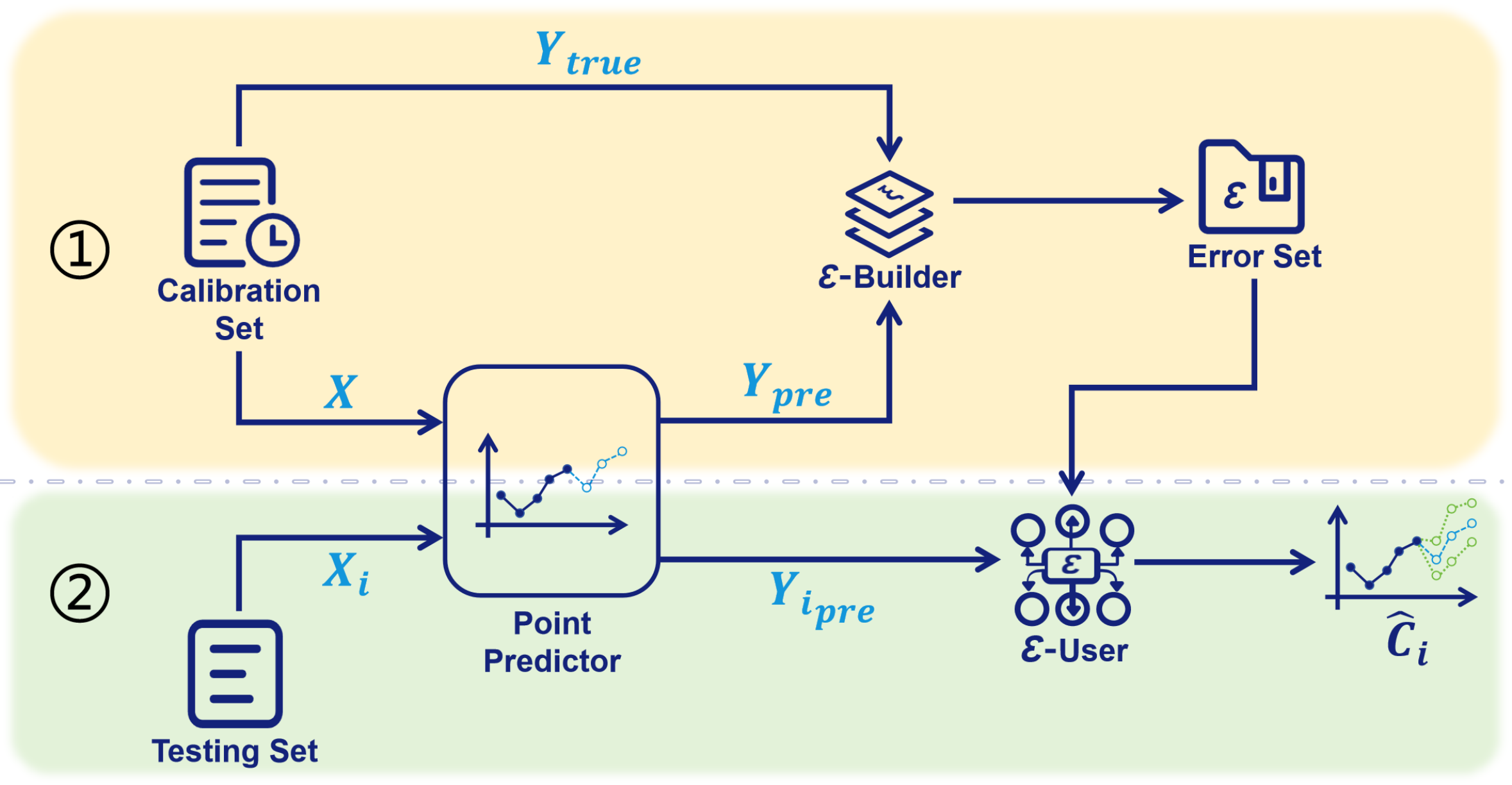}
    \caption{The workflow of CP in time-series forecasting, illustrating the preparation of error sets \(\mathcal{E}\) from calibration data and the construction of prediction intervals \(\hat{C_i}\) for test data using quantiles of \(\xi\).}
    \label{fig:Traditional CP structure}
\end{figure}

The CP workflow in time-series is illustrated in Fig.~\ref{fig:Traditional CP structure}. In the preparation stage 1, the input data \(X\) from the calibration set is first fed into the point predictor to generate the prediction results \(Y_{\text{pre}}\). The point predictor can be any predictive model that has been trained to provide point prediction results.
Subsequently, \(Y_{\text{pre}}\) are compared with the true results \(Y_{\text{true}}\) in the calibration set and passed into the \(\mathcal{E}\)-Builder to calculate the error term \(\xi\) with Eq.~\eqref{eq:Absolute value for epsilon}:
\begin{equation}
\xi = |Y_{\text{true}} - Y_{\text{pre}}| ~.\label{eq:Absolute value for epsilon}
\end{equation}
All \(\xi\) values are then assembled into the error set \(\mathcal{E}\), which serves as the basis for constructing the prediction intervals.

In the use stage 2, the input data \(X_i\) from the test set is fed into the point predictor to generate prediction \(Y_{i_\text{pre}}\). Then \(Y_{i_\text{pre}}\) is provide to the \(\mathcal{E}\)-User, which constructs a prediction interval \(\hat{C_i}\) for the \(Y_{i_\text{pre}}\). The \(\mathcal{E}\)-User determines the upper and lower bounds of the prediction interval by selecting the \(\xi\) value at the \(\alpha/2\) quantile as the lower error value and the \((1-\alpha/2)\) quantile as the upper error value from \(\mathcal{E}\). By adding these error values to \(Y_{i_\text{pre}}\), it calculates the lower and upper bounds of the \(\hat{C_i}\). Here, \(\alpha \in [0, 1]\) determines that the model has \(1-\alpha\) probability of the \(\hat{C_i}\) covering the \(Y_{i_{true}}\). The parameter \(\alpha\) is a user-defined hyperparameter, set based on practical requirements, and directly controls the trade-off between reliability and interval width.

When applied directly to time series data, CP faces certain challenges. Time series data typically exhibit significant autocorrelation and dynamic change characteristics, which CP fails to fully consider. This may lead to insufficient coverage or excessively wide prediction intervals.

To address these issues, researchers have proposed a variety of improved CP variants for time series data. These methods enhance the adaptability of CP to time series data through a number of measures. The main improvement directions and structures of these improved methods are briefly summarized in the following section.

\subsection{Improved CP for Time Series}
In recent years, numerous studies have focused on enhancing CP to better suit the quantification of uncertainty in time series data. These methods can be broadly classified into four categories:

\subsubsection*{Dynamically Updating Error Set:}
\begin{figure}[htbp]
\centering
\includegraphics[width=0.8\linewidth]{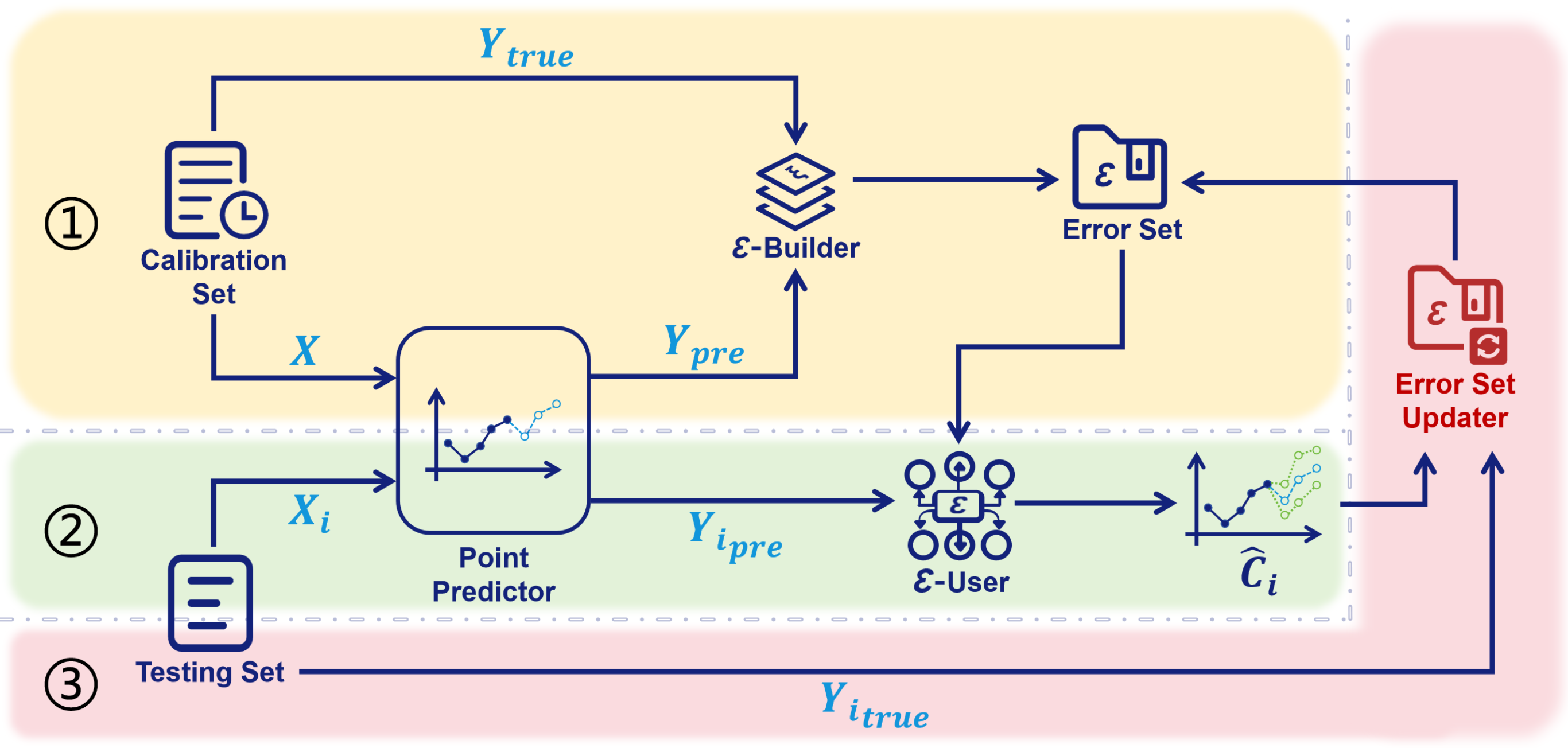}
\caption{The workflow of CP with dynamically updated error set \(\mathcal{E}\) includes a dynamic update stage where real-time errors \(\xi\) are added to \(\mathcal{E}\) using test set values \(Y_{i_\text{true}}\), thereby enhancing adaptability to time-series data.}
\label{fig:Dynamically Updating Error Set}
\end{figure}

Compared to CP, EnbPI~\cite{xu2021conformal} adds an updating stage 3, where the real-time \(\xi\) is updated into \(\mathcal{E}\) by passing the true value \(Y_{i_\text{true}}\) from the test set into the Error Set Updater along with the corresponding \(\hat{C_i}\). Such a dynamically updated \(\mathcal{E}\) over time enables EnbPI to better adapt to the characteristics of time series data. Fig.~\ref{fig:Dynamically Updating Error Set} illustrates the simplified workflow of this category method.

NexCP~\cite{barber2023conformal} adapts to the characteristics of the time series data by weighting the \(\xi\), considering recently acquired \(\xi\) as more informative. This approach approximates the effect of updating the error set, thus categorizing NexCP within this category of improved methods.

\subsubsection*{Dynamic Adjustments to \(\alpha\):}  
The core idea of this improvement method is to introduce an \(\alpha\)-Updater in the updating stage, which dynamically adjusts \(\alpha\) using \(\alpha_{t+1} = \alpha_{t} + \gamma(\alpha - err_t)\). Here, \(\gamma > 0\) is a fixed step size parameter, and \(err_t = 1\) if \(Y_t \notin \hat{C}_t(\alpha_t)\); otherwise, \(err_t = 0\).

\begin{figure}[htbp]
\centering
\includegraphics[width=0.8\linewidth]{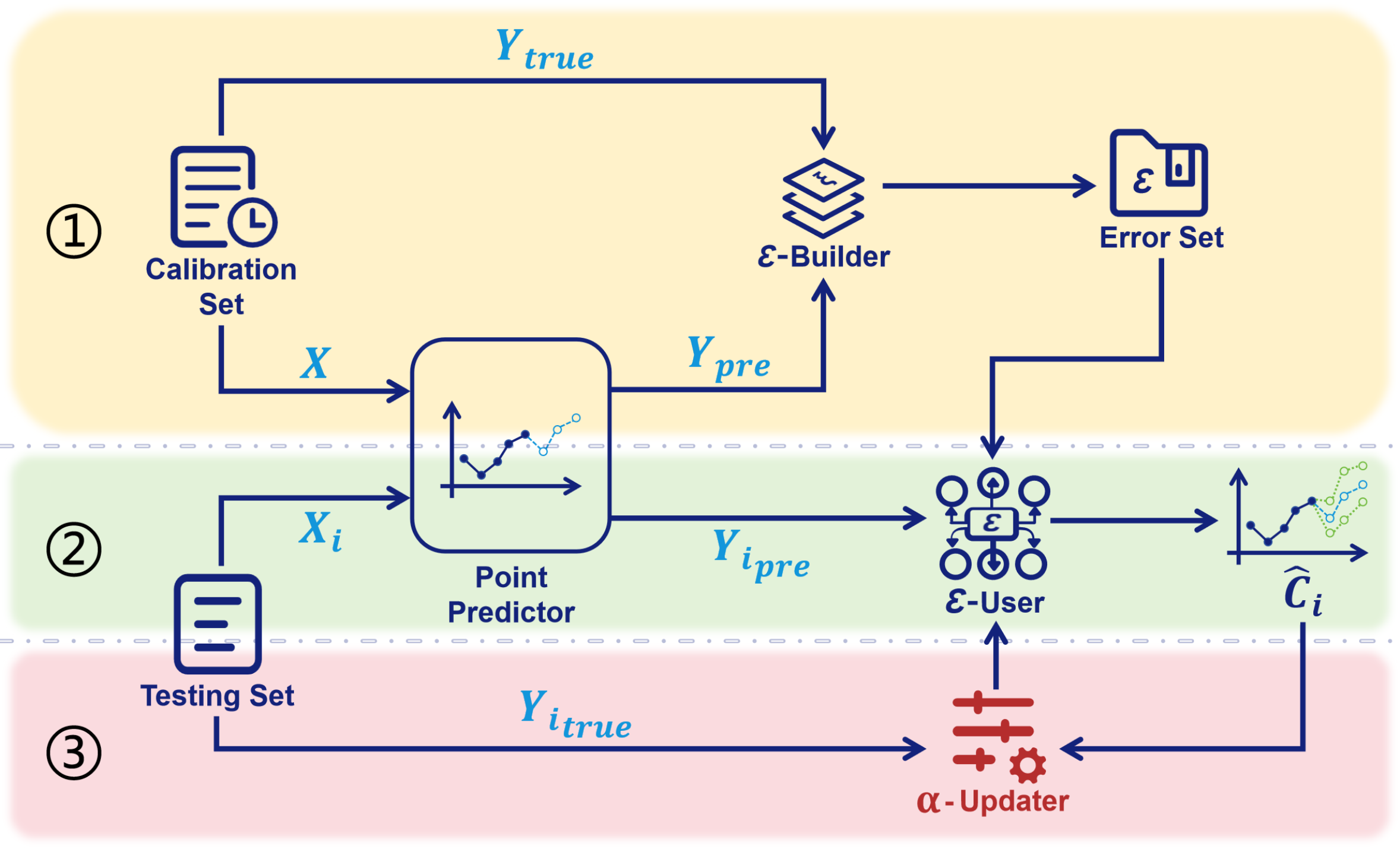}
\caption{The workflow of CP with dynamic \(\alpha\) adjustment, where the \(\alpha\)-Updater modifies the confidence level \(\alpha\) based on recent prediction errors \(err_t\), enabling adaptive prediction intervals for time-series data.}
\label{fig:Dynamic Adjustments to α}
\end{figure}

Fig.~\ref{fig:Dynamic Adjustments to α} illustrates a simplified framework of this category, where the forecast intervals are dynamically adjusted based on recent prediction performance, helping to capture changes in the distribution of time-series data.
This category includes several methods, such as ACI~\cite{gibbs2021adaptive}, CHR~\cite{sesia2021conformal}, MVP~\cite{bastani2022practical}, TQA~\cite{lin2022conformal}, and PID~\cite{angelopoulos2024conformal}.

\begin{figure}[htbp]
\centering
\includegraphics[width=0.8\linewidth]{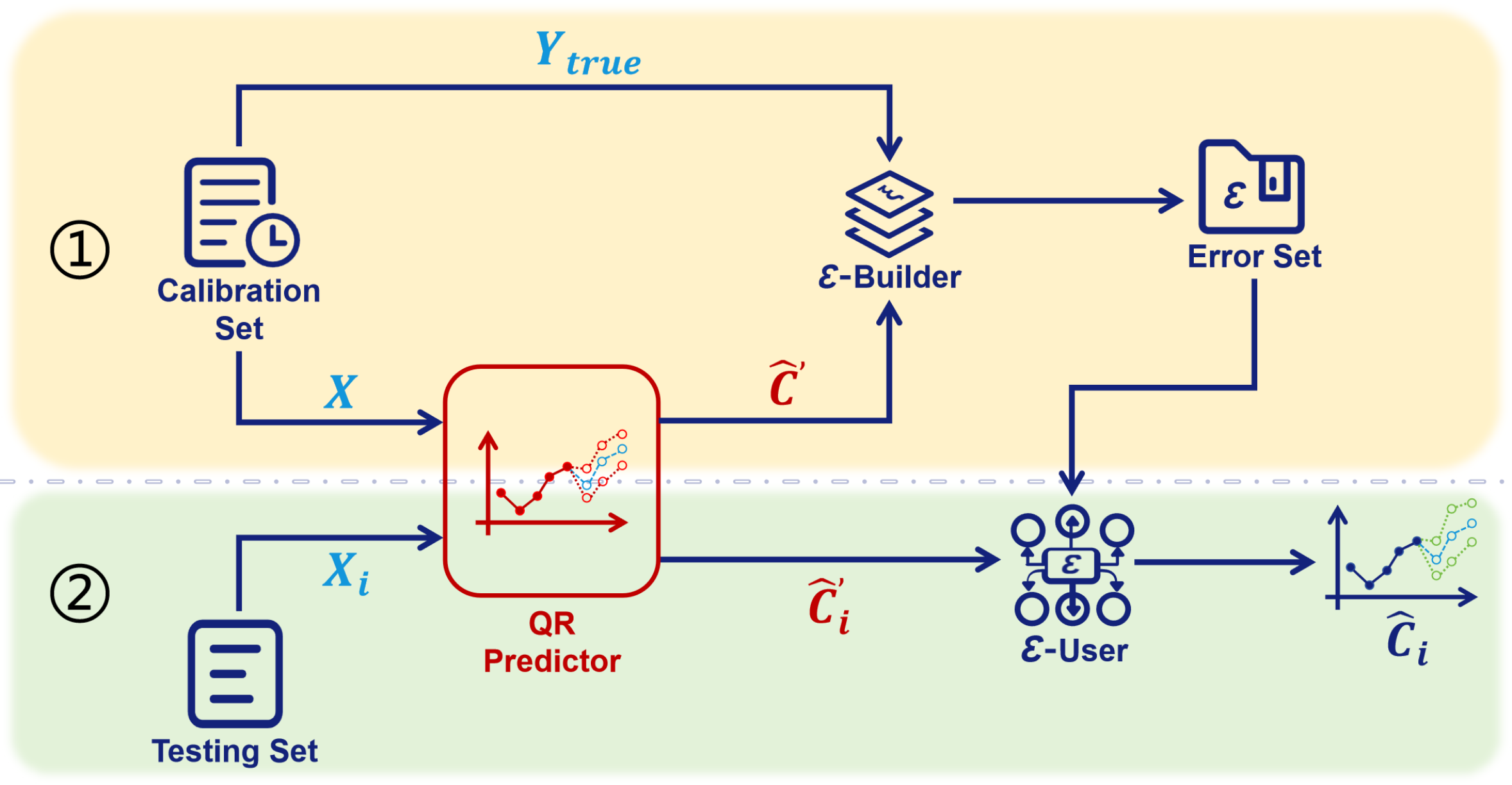}
\caption{The workflow of CP with introduce QR, where the point predictor is replaced by a QR predictor to generate preliminary bounds \(\hat{C}^{'}\), which are then used to construct the final prediction intervals \(\hat{C_i}\).}
\label{fig:Introduces Quantile Regression}
\end{figure}

\subsubsection*{Introduces Quantile Regression:}
This type of improved method replaces the point predictor in CP with a quantile regression (QR)~\cite{koenker1978regression} predictor, which constructs a preliminary \(\hat{C}^{'}\) by predicting the \((1-\alpha/2)\) and \(\alpha/2\) quantile values of the dataset to be predicted as its lower and upper bounds.
Unlike CP, this method replaces \(Y_{\text{pre}}\) and \(Y_{i_\text{pre}}\) with the upper and lower bounds in \(\hat{C}^{'}\), which are then passed into the \(\mathcal{E}\)-Builder and \(\mathcal{E}\)-User for subsequent operations to obtain the final \(\hat{C_i}\). 
Fig.~\ref{fig:Introduces Quantile Regression}. illustrates the simplified workflow of this category method, which includes approaches such as SPCI~\cite{xu2023sequential} and CQR~\cite{romano2019conformalized}.

\begin{table}[h!]
\centering
\renewcommand{\arraystretch}{1.7}
\setlength{\tabcolsep}{5pt}
\caption{Comparative analysis of model capabilities for time series forecasting, evaluating suitability for data scenarios such as variance fluctuation and small sample size, along with key features including black box compatibility and multi-step forecasting. DSCP is highlighted for its unique multi-step forecasting capability.}
\begin{adjustbox}{width=1\textwidth}
\begin{tabular}{|>{\raggedright\arraybackslash}m{2.7cm}|>{\raggedright\arraybackslash}m{4.5cm}|c|c|c|c|c|}
\hline
\multicolumn{2}{|c|}{\textbf{\begin{tabular}[c]{@{}c@{}} Comparative Analysis of Model Capabilities\end{tabular}}} & \textbf{\begin{tabular}[c]{@{}c@{}}Dynamically Updating \\ Error Set\end{tabular}} & \textbf{\begin{tabular}[c]{@{}c@{}}Dynamic \\ Adjustments $\alpha$ \end{tabular}} & \textbf{Introduces QR} & \textbf{\begin{tabular}[c]{@{}c@{}}Horizontal Slice \\ Error Set\end{tabular}} & \textbf{DSCP} \\ \hline
\multirow{2}{*}{\textbf{Data Scenarios}} & Variance Fluctuation & \checked & \checked & \checked & \checked & \checked \\ \cline{2-7} 
 & Small Sample Size & \checked & \checked & × & × & × \\ \hline
\multirow{4}{*}{\textbf{Model Features}} & Black Box & \checked & \checked & × & \checked & \checked \\ \cline{2-7} 
 & None Additional Model & \checked & \checked & × & × & \checked \\ \cline{2-7} 
 & Enable Update & \checked & \checked & × & \checked & \checked \\ \cline{2-7} 
 & \textbf{Multi-step Forecasting} & \textbf{\ding{55}} & \textbf{\ding{55}} & \textbf{\ding{55}} & \textbf{\ding{55}} & \textbf{\checkmark} \\ \hline
\end{tabular}
\end{adjustbox}
\label{table:model_capabilities}
\end{table}

\subsubsection*{Horizontally Split Error Set:}
The main idea of this improved method is to introduce an \(\mathcal{E}\)-selector in use stage 2. It compares the temporal features of \(Y_{i_\text{pre}}\) and all \(\xi\) in \(\mathcal{E}\), selectively extracts a subset of errors \(\mathcal{E}'\), that is most relevant to the current \(Y_{i_\text{pre}}\). The \(\mathcal{E}\)-User then uses the \(\mathcal{E}'\) instead of the \(\mathcal{E}\) to construct \(\hat{C_i}\). By dynamically extracting the \(\mathcal{E}'\) most relevant to the current moment from \(\mathcal{E}\), the method can better adapt to changes in the distribution of time series data.

\begin{figure}[htbp]
\centering
\includegraphics[width=0.8\linewidth]{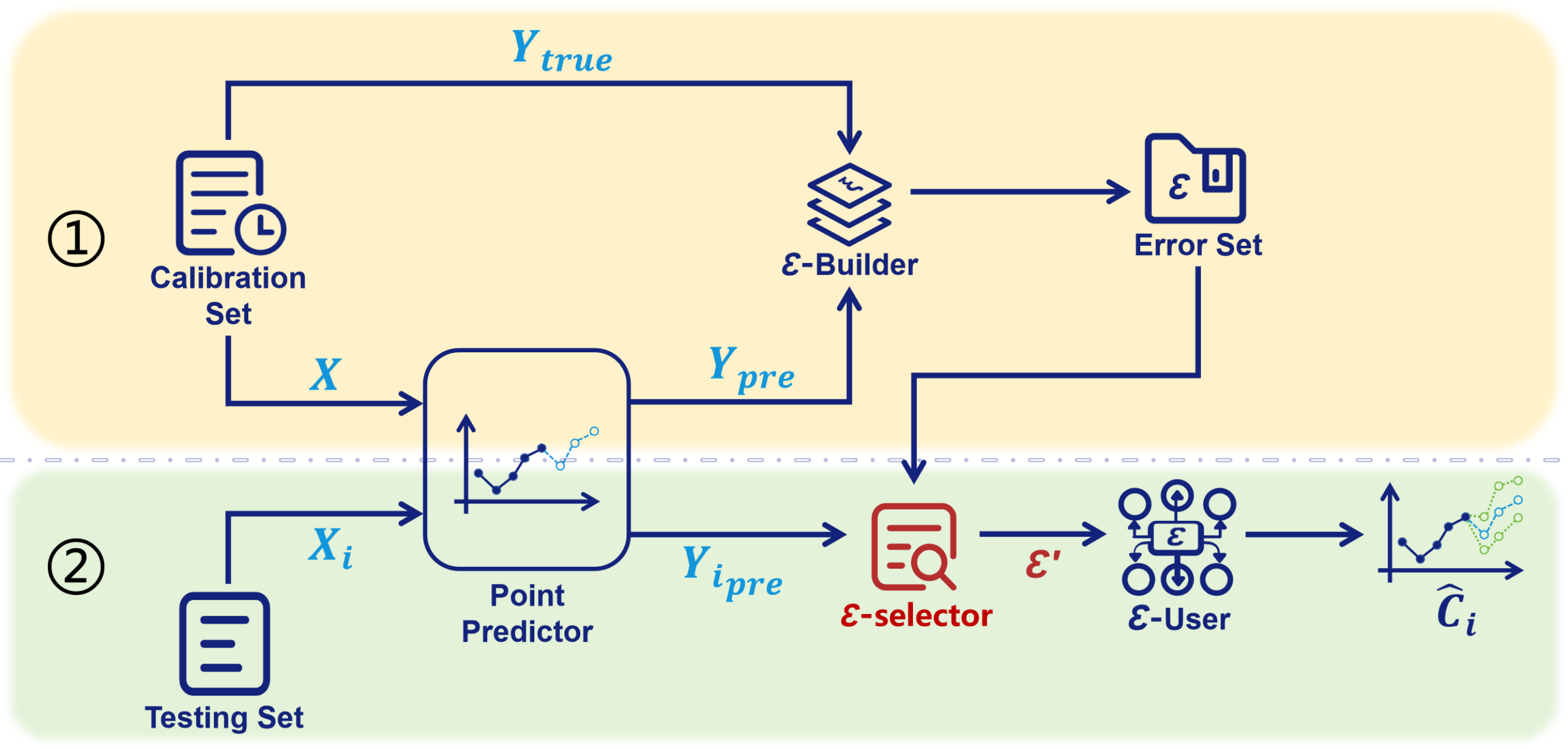}
\caption{The workflow of CP methods with horizontally split error sets, where an \(\mathcal{E}\)-selector dynamically extracts a subset \(\mathcal{E}'\) based on temporal features of \(Y_{i_\text{pre}}\), enabling adaptive prediction intervals \(\hat{C_i}\) for time-series data.}
\label{fig:Horizontally Slice Error Set}
\end{figure}

Fig.~\ref{fig:Horizontally Slice Error Set}. illustrates the simplified workflow of this category method, which includes approaches such as CF-RNN~\cite{stankeviciute2021conformal}, CopulaCPTs~\cite{sun2022copula} and HopCPT~\cite{auer2024conformal}.

\hyperref[table:model_capabilities]{Table~\ref*{table:model_capabilities}} summarizes the data scenarios suitable for each improved CP variants and highlights their key features, including the DSCP introduced in this paper. The table evaluates methods based on their compatibility with data scenarios such as variance fluctuation and small sample size, as well as key features like black box compatibility and multi-step forecasting capability. Notably, DSCP stands out for its unique ability to handle multi-step forecasting, addressing the limitations of other methods in this context. This capability is particularly relevant to the issues discussed in this paper, where multi-step forecasting is essential for applications like resource scheduling and risk management.

\subsection{Limitations of CP in Time Series Forecasting}
Although these four types of improved CP variants have optimized CP for time series forecasting, they still exhibit notable limitations in their design and practical application, particularly in handling complex data distributions and dynamic uncertainty variations:
\begin{itemize}[leftmargin=*, itemsep=2pt, parsep=1pt]
    \item \textbf{Mixed Use of \(\xi\) Across Data Subsets}: Datasets often consist of several data subsets with different data distributions. And most improved CP variants have used error term \(\xi\) from different data subsets together.
    \item \textbf{Failure to Address Uncertainty Variations Within Time Windows}: Most improved CP variants fail to account for the varying uncertainties across different time steps within a prediction, as they treat the entire prediction as a single window. In our approach, a window is defined as a collection of time steps where \(\xi\) values are shared collectively. While some improved CP variants~\cite{stankeviciute2021conformal,stankeviciute2021conformal,auer2024conformal} attempt to address this issue by treating each time step as a single window, they ignore the potential consistency in uncertainty distributions between adjacent time steps. This approach unnecessarily reduces the amount of statistical data within each window, potentially degrading the performance of CP as a statistical method.
\end{itemize}

\begin{figure*}[htbp] 
\centering 
\includegraphics[width=1\textwidth]{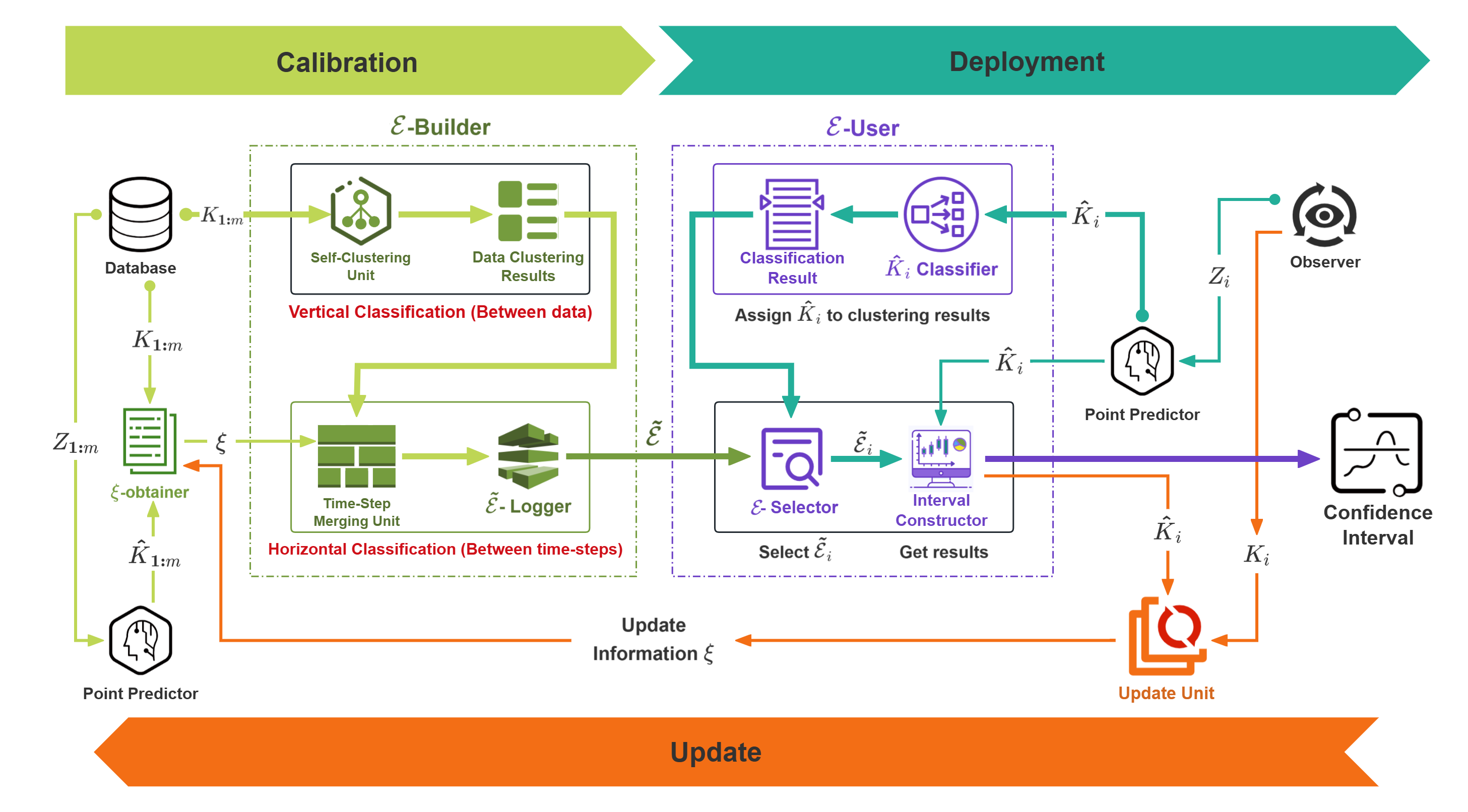}
\caption{Workflow of the DSCP, divided into three stages: calibration, deployment, and update. In the calibration stage, the \(\mathcal{E}\)-Builder module first vertically splits the calibration set of historical predictions \(K_{1:m}\) using a self-clustering unit to obtain partitioned by class. Then, within each class, it performs horizontal splitting on the intra-class \(\xi\)-obtainer information to dynamically partition \(\xi\) across time steps. The processed results, after class-wise partitioning and dynamic merging of time steps, are stored in the \(\tilde{\mathcal{E}}\)-Logger. In the deployment stage, the \(\hat{K}_i\) Classifier assigns new input data \(\hat{K}_i\) to the appropriate cluster. Then, the \(\mathcal{E}\)-Selector retrieves information from the \(\tilde{\mathcal{E}}\)-Logger to construct prediction intervals for the prediction \(\hat{K}_i\) based on the clustering results. In the updating stage, the \(\hat{K}_i\) is compared with actual data \(K_i\) to update the \(\xi\)-obtainer.} 
\label{fig:structure} 
\end{figure*}

\section{Dual-Splitting Conformal Prediction}
To address the aforementioned limitations, we propose DSCP, a cohesive approach for reliable multi-step time-series forecasting. The overall workflow of DSCP, illustrated in Figure~\ref{fig:structure}, is divided into three stages: calibration, deployment, and updating. In the calibration stage, the model processes the calibration set to construct error sets \(\mathcal{E}\). In the deployment stage, it assigns new prediction \(\hat{K}_i\) to clusters and constructs prediction intervals \(\hat{C}\). In the updating stage, the model computes the error \(\xi\) by comparing the predicted value \(\hat{K}_i\) with the real result \(K_i\), and then uses this error \(\xi\) to refine the \(\mathcal{E}\). This three-stage workflow enables DSCP to handle multi-step forecasting effectively.

We begin by defining key concepts, including datasets, error terms \(\xi\), and prediction intervals \(\hat{C}\). Next, we present the motivation behind DSCP, focusing on vertical and horizontal classification strategies to improve the accuracy of uncertainty quantification. Finally, we describe how to effectively address the challenges outlined in the previous section and integrate these measures into the development of the DSCP framework.

\subsection{Definitions}
\textbf{Dataset and point predictor:} 
In this study, we use three separate datasets: training set, calibration set, and testing set, which are mutually exclusive. 
The training set is utilized to train the point prediction model \(f\). The calibration set is used to get error terms \(\xi\) and error set \(\mathcal{E}\). The test set is reserved for evaluating the performance of CP.

The dataset includes \(X=[x_1, x_2, \ldots, x_t]\), where each \(x_t=[\theta_1, \theta_2, \ldots, \theta_n]\) represents the \(n\) features at time \(t\). The corresponding target values are \(Y= [y_1, y_2, \ldots, y_t]\), where \(y_t\) denotes the observed value at time \(t\). 
For prediction, the input data is constructed as \(Z_t = [x_{t-a}, x_{t-a+1}, \ldots, x_t] \in \mathbb{R}^{a \times n}\), where \(a\) is the size of the input window. The model \(f\) receives the input \(Z_t\) and then produces the output for the next \(b\) time steps, denoted as \(K_t = [\hat{y}_{t+1}, \hat{y}_{t+2}, \ldots, \hat{y}_{t+b}]\).

\textbf{Error Terms, Error Set:} 
We define the error term \(\xi_t\) as the difference between true value \(y_t\) and predicted value \(\hat{y}_t\):
\begin{equation}
\xi_t = y_t - \hat{y}_t ~.\label{eq:None_Absolute value solving epsilon}
\end{equation}
The \(\mathcal{E} = \{\xi_1,\xi_2,\dots,\xi_t\}\) consists of all \(\xi_t\) values derived from the calibration set, serving as the basis for CP to estimate uncertainty.

CP and its variations~\cite{xu2021conformal,barber2023conformal,gibbs2021adaptive,bastani2022practical,lin2022conformal,angelopoulos2024conformal,stankeviciute2021conformal,sun2022copula,romano2019conformalized} use Eq.~\eqref{eq:Absolute value for epsilon} to calculate \(\xi_t\), but they treat both overestimations \(\xi_{over}\) and underestimations \(\xi_{under}\) as absolute values, thereby preventing their distinction. As a result, these two different types of uncertainty are mixed together, which can negatively affect the construction of the \(\hat{C}_i\), ultimately reducing the accuracy of \(\hat{C_i}\). As illustrated in Fig.~\ref{fig:Suitability}(b), the \(\xi\) in the central region primarily represent \(\xi_{under}\), while those at both ends predominantly correspond to \(\xi_{over}\). 

To address this issue, the DSCP method introduces Eq.~\eqref{eq:None_Absolute value solving epsilon} as a replacement for Eq.~\eqref{eq:Absolute value for epsilon}, enabling the distinction between \(\xi_{over}\) and \(\xi_{under}\) by assigning positive and negative values to them, respectively. This approach facilitates the construction of asymmetric upper and lower bounds, effectively handling both \(\xi_{over}\) and \(\xi_{under}\).

\textbf{CP's Objective and Prediction Interval:}  
CP's objective is to ensure that the prediction interval \(\hat{C}_t^\alpha(K_t)\) contains the true values \(Y_t=[y_{t+1:t+b}]\) with a probability of at least \((1 - \alpha)\), where \((1-\alpha)\) is the confidence level of the interval. This is expressed as: 
\(\text{Pr}\{Y_{t} \in \hat{C}_t^\alpha(K_t)\} \geq 1 - \alpha\).

To achieve this, we construct the prediction interval \(\hat{C}_t^\alpha(K_t)\) for the \(b\)-step predictions \(K_t\) of \(f\), which provides upper and lower bounds. The interval is defined as:
\begin{equation}
\hat{C}_t^\alpha(K_t) = 
\left\{
\begin{aligned}
    &\{\hat{y}_{t+1}, \ldots, \hat{y}_{t+b}\} + Q_{1-\alpha/2}(\mathcal{E}), \\
    &\{\hat{y}_{t+1}, \ldots, \hat{y}_{t+b}\} + Q_{\alpha/2}(\mathcal{E}),
\end{aligned}
\right.
\end{equation}
where \(Q_{1-\alpha/2}(\mathcal{E})\) and \(Q_{\alpha/2}(\mathcal{E})\) are the values corresponding to the \((1-\alpha/2)\)-th and \((\alpha/2)\)-th quantiles of \(\mathcal{E}\), respectively. The interval \([Q_{\alpha/2}(\mathcal{E}), Q_{1-\alpha/2}(\mathcal{E})]\) contains \((1-\alpha)\) of the values in \(\mathcal{E}\).

\subsection{Motivation}
The core idea of DSCP is to extract error term \(\xi\) from historical predictions \(\{K_t\}_{t=1}^m\) that are similar to new prediction \(\hat{K}_i\), and then use these \(\xi\) to construct the \(\hat{C_i}\) for \(\hat{K}_i\). This approach ensures that only relevant \(\xi\) are used for \(\hat{K}_i\), excluding unrelated \(\xi\). To achieve this, the classification is conducted along two dimensions, as below:

\textbf{Vertical Classification} groups the \(\{K_t\}_{t=1}^m\) into \(k\) classes based on their temporal trends and amplitude magnitudes. This approach is motivated by the observation that different predictions \(K_t\) often exhibit distinct patterns due to variations in underlying conditions. For example, a \(K_t\) generated under one set of conditions may exhibit a specific trend and magnitude, whereas another \(K_t\) under different conditions may display entirely different behavior. These condition-dependent variations result in distinct uncertainties embedded within \(\{K_t\}_{t=1}^m\).

By categorizing \(K_t\) based on their trends and magnitudes, we group \(\{K_t\}_{t=1}^m\) with similar underlying conditions and uncertainties. This ensures that the constructed \(\hat{C}_i\) are based on consistent \(\xi_t\), leading to more reliable uncertainty quantification. Various methods can be employed for vertical classification, such as k-Nearest Neighbors~\cite{peterson2009k}, Dynamic Time Warping~\cite{muller2007dynamic}, Hidden Markov Models~\cite{rabiner1986introduction}, Convolutional Neural Networks~\cite{yamashita2018convolutional}, and Long Short-Term Memory Networks~\cite{graves2012long}. In this work, we employ the k-means approach~\cite{ahmed2020k} to group \(\{K_t\}_{t=1}^m\) into distinct categories.

\textbf{Horizontal Classification} groups time steps within a category \(K_t\) into several windows based on their \(\xi\) distributions. Even within a single category of \(K_t\), the \(\xi\) distributions can vary significantly across different time steps due to temporal dependencies and dynamic data characteristics. For instance, a time step in the morning may exhibit a different \(\xi\) distribution compared to a time step in the night, even if both belong to the same category of \(K_t\).

To address this variability, we partition the time steps within each category into multiple windows, ensuring that the \(\xi\) distributions within each window are consistent. The \(\xi\) collected at a specific time step is only shared within its corresponding window, preventing the mixing of error terms from different \(\xi\) distributions. This ensures that the constructed \(\hat{C}_i\) are accurate and reliable.

In summary, motivated by these two observations, the core idea of DSCP is to perform dual-dimensional classification of error terms \(\xi_t\). Vertical classification groups historical predictions \(\{K_t\}_{t=1}^m\) based on their temporal trends and amplitude magnitudes, capturing the underlying patterns in the data, while horizontal classification partitions time steps within each category based on their \(\xi\) distributions, ensuring that each partition reflects consistent uncertainty patterns. This approach ensures that the constructed \(\hat{C}_i\) are based on consistent, contextually relevant, and statistically robust error information, significantly improving the accuracy, reliability, and interpretability of uncertainty quantification in multi-step time series forecasting.

\begin{figure}[htbp]
\centering
\includegraphics[width=1\linewidth]{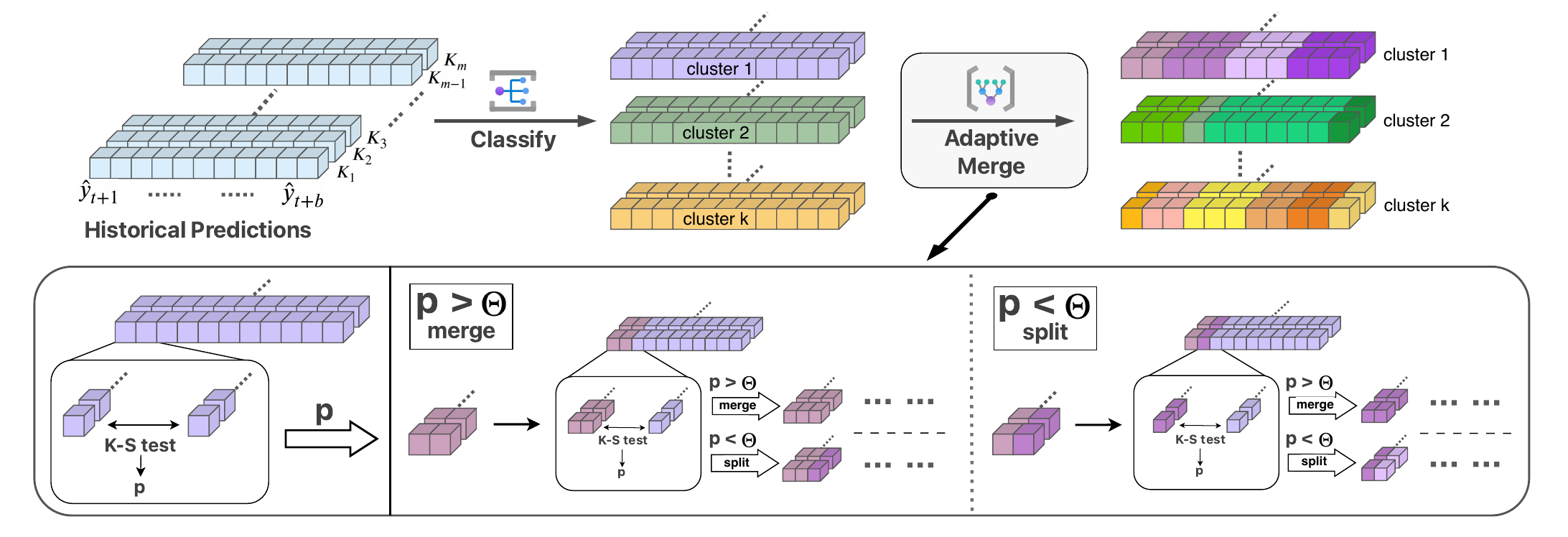}
\caption{The figure illustrates the two-stage process of DSCP during its calibration phase. In the first stage, historical predictions \(\{K_t\}^m_{t=1}\) are grouped into clusters 1 to \(k\) using a suitable classification method. Predictions within the same cluster are assigned the same color. In the second stage, the K-S test is used to evaluate the similarity of \(\xi\) distributions between adjacent time steps within each cluster. If the p-value exceeds a predefined threshold \(\Theta\), the time steps are merged into the same window. Otherwise, they are divided into different windows. Here, \(\Theta\) is a user-defined hyperparameter that balances the trade-off between the amount of data sampled within the window and the degree of consistency in the \(\xi\) distribution. Time steps within the same window are assigned the same color, while different windows within the same cluster are represented by shades of the same color family.}
\label{fig:Adaptive merge}
\end{figure}

\subsection{Workflow}
To address the challenges of time-varying uncertainties and diverse scenarios in time series data, DSCP employs a two-stage process: \textbf{clustering historical predictions} and \textbf{adaptively merging time steps}, as illustrated in Figure~\ref{fig:Adaptive merge}. These two stages are performed during the preparation phase on the calibration set, and the resulting classification is later used in the testing phase to select appropriate \(\xi\) subsets for constructing confidence intervals \(\hat{C}_i\) for new prediction \(\hat{K}_i\).

First, historical predictions \(\{K_t\}^m_{t=1}\) are grouped into clusters based on their similarity using a suitable classification method. In this work, we adopt k-means approach~\cite{ahmed2020k} to ensure that \(K_t\) with similar trends and magnitudes are grouped together. This step results in \(k\) clusters of \(\{K_t\}^m_{t=1}\), each representing a distinct group of \(K_t\) with shared characteristics. In Figure~\ref{fig:Adaptive merge}, predictions within the same cluster are assigned the same color, visually distinguishing them from predictions in other clusters. By clustering the \(\{K_t\}^m_{t=1}\), we can effectively handle diverse scenarios in the data and ensure that subsequent error analysis is conducted within consistent groups.

Next, the Kolmogorov-Smirnov (K-S) test~\cite{berger2014kolmogorov} is utilized to evaluate the similarity of \(\xi\) distributions between adjacent time steps in a cluster. The K-S test produces a p-value that quantifies the consistency of \(\xi\) distributions between two adjacent time steps, providing a statistical measure of their similarity. If the p-value exceeds a predefined threshold \(\Theta\), the time steps are merged into the same window. Otherwise, they are divided into different windows. In Figure~\ref{fig:Adaptive merge}, time steps within the same window are assigned the same color, while different windows within the same cluster are represented by shades of the same color family.

\begin{figure}[htbp]
\centering
\includegraphics[width=0.8\linewidth]{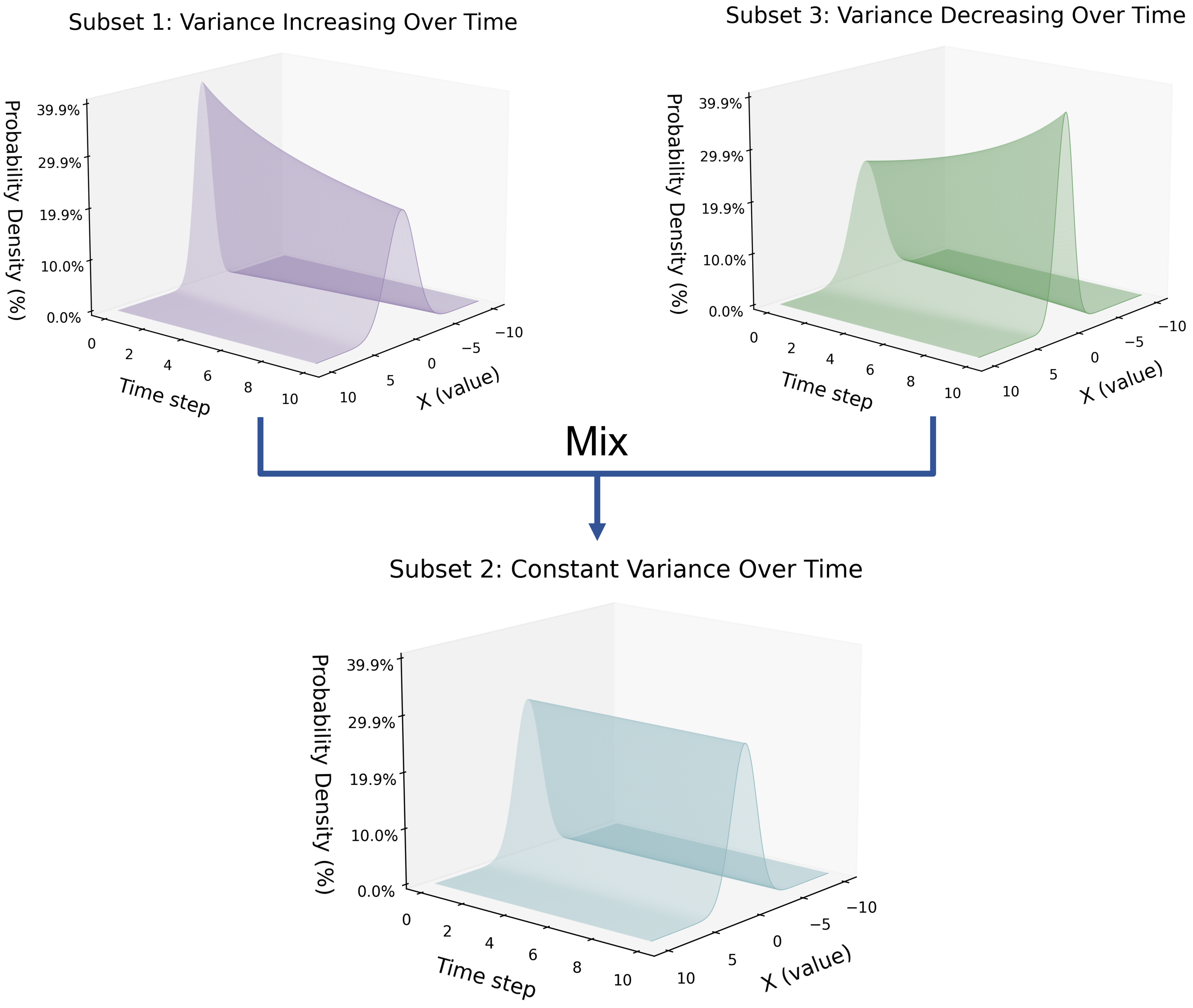}
\caption{Schematic representation of the mixing phenomenon in probability density functions. The x-axis represents the probability distribution values, the z-axis denotes Probability Density indicating the likelihood of these values, and the y-axis represents Time showing the temporal dimension. This figure illustrates how merging data subsets with time-varying variances may create the illusion of a dataset with constant overall variance.}
\label{fig:Data Consolidation}
\end{figure}

\subsubsection*{Clustering Prediction Results Based on Similarity}
Datasets frequently comprise multiple subsets with diverse data distributions. As illustrated in Fig.~\ref{fig:Data Consolidation}, time-varying data distribution is hidden behind the seemingly stationary overall distribution. For instance, in solar energy datasets, data distributions vary significantly between sunny, cloudy, and rainy days. These varying conditions result in distinct data distribution characteristics. If the dataset subsets are not properly partitioned, the error term \(\xi\) across different subsets may interfere with each other, thereby weakening the accuracy and effectiveness of CP.

To address this issue, DSCP employs a self-clustering method \(\psi\) to classify \(\{K_t\}_{t=1}^m\) into \(k\) clusters, defind as:
\begin{equation}
G_{K_1}, \ldots, G_{K_m} \leftarrow \psi(\{K_{t}\}_{t=1}^m,~N)
\end{equation}
where \(G_{K_t}\) represents the cluster assigned to \(K_t\), with values from 1 to \(k\). The self-clustering method \(\psi\) employs the k-means algorithm~\cite{macqueen1967some}, testing different values of \(k\) from 1 to the maximum number \(N\). It then evaluates all \(N\) clustering results using silhouette scores~\cite{rousseeuw1987silhouettes}. The silhouette score is used to measure the quality of the clustering result, with higher scores indicating better clustering. The clustering result with the highest silhouette score is selected as the final output \(G_{K_1}, \ldots, G_{K_m}\).

When a prediction \(\hat{K_i}\) is obtained from the test set, we calculate the similarity scores \(e_1, e_2, \dots, e_m\) between \(\hat{K_i}\) and the historical predictions \(\{K_t\}_{t=1}^m\) using the soft-DTW method~\cite{cuturi2017soft}, as illustrated by the \(\hat{K_i}\) Classifier in Figure~\ref{fig:structure}. These similarity scores are sorted in descending order, and the largest \(s\) values are selected, where \(s\) is the size of the smallest cluster. The indices corresponding to these \(s\) values are stored in the set \(\text{I}\), and the corresponding historical predictions \(K_t\) are referred to as the selected \(K_t\).

Next, we determine the cluster assignment \(G_{\hat{K_i}}\) for \(\hat{K_i}\) by analyzing the clusters of the selected \(K_t\). Specifically, we calculate the frequency of each cluster index \(x\) among the clusters of the selected \(K_t\) and assign \(\hat{K_i}\) to the cluster index \(x\) with the highest frequency.

To construct the error subset \(\mathcal{E}_i'\), we collect all error terms \(\xi_t\) from the \(K_t\) that belong to the cluster \(G_{\hat{K_i}}\). Specifically, \(\mathcal{E}_i'\) is defined as:\(\mathcal{E}_i' = \bigcup_{t \in T_{\hat{K_i}}} \{\xi_{t,1}, \xi_{t,2}, \ldots, \xi_{t,b}\},\)
where \(T_{\hat{K_i}}\) is the set of indices \(t\) for which \(G_{K_t} = G_{\hat{K_i}}\). Here, \(t\) refers to the index of \(K_t\) in the historical predictions \(\{K_t\}_{t=1}^m\), and the subscript \(1, 2, \ldots, b\) indicates the time steps within the prediction window of size \(b\) for each \(K_t\). The error terms \(\xi_{t,1}, \xi_{t,2}, \ldots, \xi_{t,b}\) are obtained from these time steps. The subset \(\mathcal{E}_i'\) is then used to construct the prediction interval \(\hat{C_i}\) for \(\hat{K_i}\), as depicted in the \(\mathcal{E}\)-Selector of Figure~\ref{fig:structure}.

\begin{figure}[htbp]
\centering
\includegraphics[width=1\linewidth]{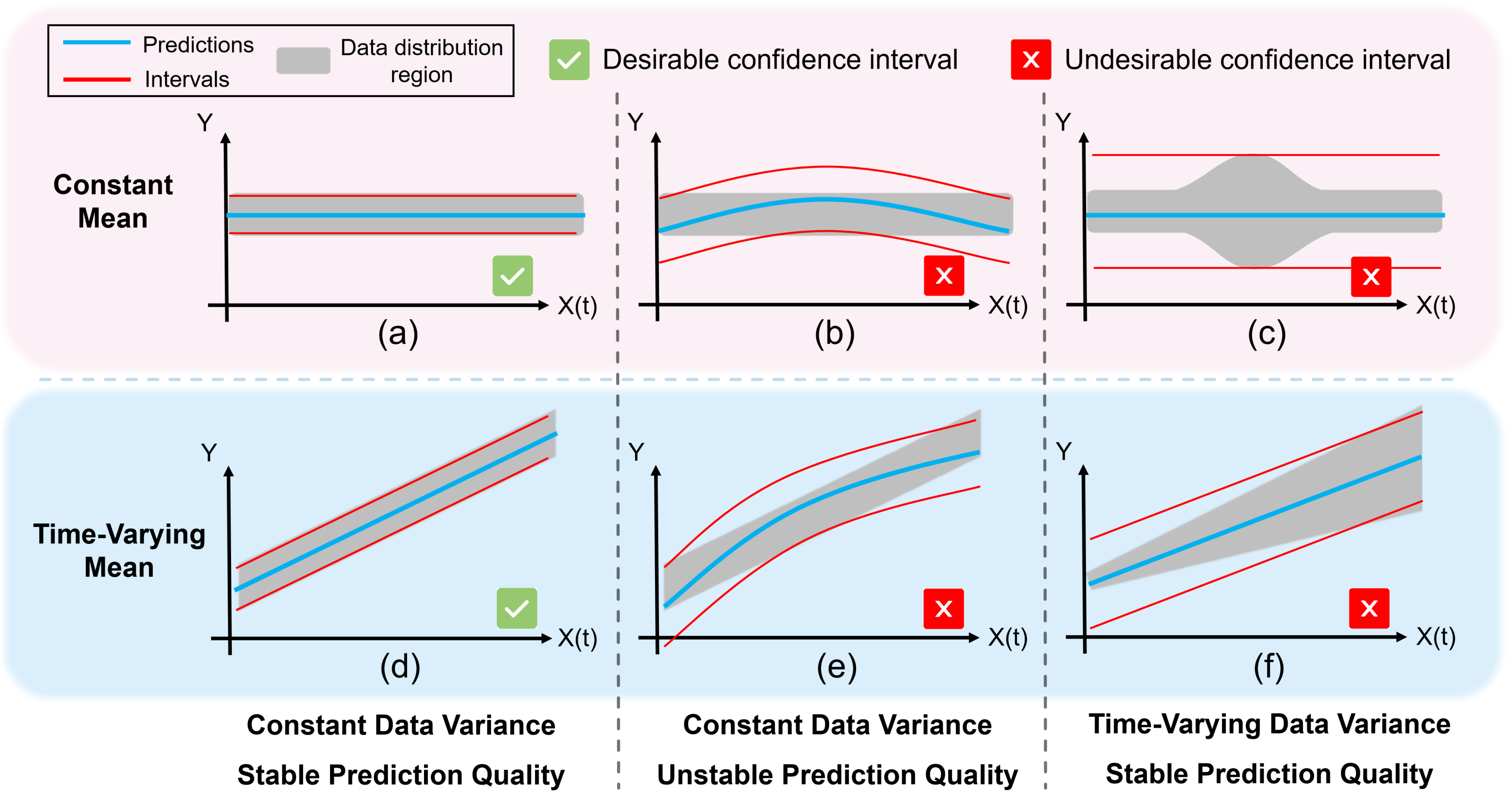} 
\caption{Confidence interval performance across scenarios. The top red region shows Stationary Mean Distribution; the bottom blue region shows Non-Stationary Mean Distribution. The figure highlights CP performance changes with variance stability, mean behavior, and prediction quality.}
\label{fig:Suitability}
\end{figure}

\subsubsection*{Adaptive Merging Prediction Errors Across Time Steps}
In a piece of data \(K_t\) that spans multiple time steps, uncertainties often vary significantly. For instance, in a solar energy dataset, the uncertainty differs between nighttime and daytime data. At night, solar energy generation exhibits almost no uncertainty, while during the day, the uncertainty is considerably higher. If time steps with varying uncertainties are not segmented, the differing \(\xi\) distributions may interfere with each other, thereby compromising the accuracy and effectiveness of \(\hat{C}\).

To address this issue, we propose a solution that groups time steps with similar \(\xi\) distributions into separate windows. Formally, this solution is represented by a function \(\mathcal{M}\) that takes a similarity threshold \(\Theta\) and a sequence of error subsets \(\mathcal{E}'_{(1)}, \ldots, \mathcal{E}'_{(b)}\) as input, and outputs merged error subsets:

\begin{equation}
\{\tilde{\mathcal{E}}_{(1)}, \ldots, \tilde{\mathcal{E}}_{(b)}\} = \mathcal{M}(\Theta, \{\mathcal{E}'_{(1)}, \ldots, \mathcal{E}'_{(b)}\}),
\end{equation}
where \(\{\mathcal{E}'_{(1)}, \ldots, \mathcal{E}'_{(b)}\}\) are the initial error subsets at each time step, assumed to be independent. The function \(\mathcal{M}\) dynamically merges these subsets based on the similarity of their \(\xi\) distributions. For example, if the first five time steps have similar distributions, their merged subsets \(\tilde{\mathcal{E}}_{(1)}, \ldots, \tilde{\mathcal{E}}_{(5)}\) will be identical. This ensures that time steps with consistent \(\xi\) distributions are grouped together, thereby preventing interference between dissimilar distributions.

The core of this solution is a dynamic merging process that iteratively evaluates and merges adjacent time steps based on their \(\xi\) distribution similarity. This approach achieves the desired segmentation while preserving sufficient data within each window for robust statistical analysis.

\textbf{Criterion for Merging:} To implement this dynamic merging process, we establish a merging criterion based on ensuring consistent \(\xi\) distributions within each window. This consistency is crucial because it directly impacts the performance of \(\hat{C}\). To illustrate this, consider the scenarios depicted in Fig.~\ref{fig:Suitability}. The figure compares the performance of \(\hat{C}\) under different conditions, focusing on two main scenarios: stationary mean distribution (top red region) and non-stationary mean distribution (bottom blue region). Within each scenario, we further analyze three cases based on data variance stability and prediction quality:
\begin{enumerate}[leftmargin=*, itemsep=0pt, parsep=1pt]
\item Constant data variance with stable prediction quality (subfigures a and d): In these cases, \(\hat{C}\) performs well, producing narrow and accurate confidence intervals.
\item Constant data variance with unstable prediction quality (subfigures b and e): In these cases, \(\hat{C}\) tends to produce excessively wide confidence intervals, often covering regions without data.
\item Time-varying data variance with stable prediction quality (subfigures c and f): Here, \(\hat{C}\) also underperforms, as the varying variance leads to inconsistent \(\xi\) distributions.
\end{enumerate}

The case of time-varying data variance with unstable prediction quality is not explicitly analyzed in Fig.~\ref{fig:Suitability}. This is because both conditions, time-varying variance and unstable prediction quality, individually lead to suboptimal performance of \(\hat{C}\), as demonstrated in subfigures (b), (e), (c), and (f). Combining these two conditions would further exacerbate the issues, resulting in confidence intervals that are both excessively wide and inconsistent. Therefore, this scenario is omitted from the analysis, as it does not provide additional insights beyond the already observed limitations.

From these observations, we conclude that \textbf{stable prediction quality} and \textbf{constant data variance} are critical for achieving reliable \(\hat{C}\). These two factors can be unified into a single criterion: the \(\xi\) distributions within a window must remain consistent. This consistency is quantitatively defined as the probability that the \(\xi\) distributions at all time steps within the window are drawn from the same distribution exceeding a threshold \(\Theta\). By ensuring this consistency, we simultaneously satisfy the conditions for stable prediction quality and constant data variance, leading to reliable \(\hat{C}\).

\begin{algorithm}
\caption{Adaptive Error Set Merging \(\mathcal{M}\)}
\label{Algorithm:Adaptive Error Set Merging}
\begin{algorithmic}[1]
\setstretch{1.1}
\State \textbf{Input:} similarity threshold \( \Theta \); each-step error sets \( \mathcal{E}'_{(1)}, \ldots, \mathcal{E}'_{(b)} \)
\State \textbf{Output:} merged error subsets \({\tilde{\mathcal{E}}_{(1)}}, \ldots, {\tilde{\mathcal{E}}_{(b)}}\)
\State Initialize $j = 1$, $m = 1$
\While{$j < b-1$}
    \State\(p_{j,j+1} = KS(\mathcal{E}'_{(j)}, \mathcal{E}'_{(j+1)})\)
    \If{$p_{j,j+1} > \Theta$}
        \State Merge \(\mathcal{E}'_{(j+1)} \gets \mathcal{E}'_{(j)} \cup \mathcal{E}'_{(j+1)}\)
        \State Increment \(j \gets j + 1\)
        \If{$j=b$}
            \For{$i = m$ to $j$}
                \State Save merged error subset \(\tilde{\mathcal{E}}_{(i)} \gets \mathcal{E}'_{(j)}\)
            \EndFor
        \EndIf
    \Else
        \For{$i = m$ to $j$}
            \State Save merged error subset \(\tilde{\mathcal{E}}_{(i)} \gets \mathcal{E}'_{(j)}\)
        \EndFor
        \State Increment \(j \gets j + 1\)
        \State Update index \(m \gets j\)
        \If{$j=b$}
            \State Save merged error subset \(\tilde{\mathcal{E}}_{(b)} \gets \mathcal{E}'_{(j)}\)
        \EndIf
    \EndIf
\EndWhile
\State \textbf{Return:} merged error subsets \({\tilde{\mathcal{E}}_{(1)}}, \ldots, {\tilde{\mathcal{E}}_{(b)}}\)
\end{algorithmic}
\end{algorithm}

Based on the merging criterion, we introduce the Adaptive Error Set Merging method, represented by the function \(\mathcal{M}\). This method processes the error subsets \(\mathcal{E}'_{(1)}, \ldots, \mathcal{E}'_{(b)}\), evaluates the similarity between \(\mathcal{E}'_{(j)}\) and \(\mathcal{E}'_{(j+1)}\) at neighboring time steps using the Kolmogorov-Smirnov (KS) test~\cite{berger2014kolmogorov}, and dynamically merges them to obtain the merged error subsets \(\tilde{\mathcal{E}}_{(1)}, \ldots, \tilde{\mathcal{E}}_{(b)}\). The KS test compares the empirical cumulative distribution functions of the two sets and generates a \(p\)-value \(p_{j,j+1}\), which quantifies the probability that \(\mathcal{E}'_{(j)}\) and \(\mathcal{E}'_{(j+1)}\) are drawn from the same distribution. If \(p_{j,j+1} > \Theta\), the two sets are merged, indicating their distributions are statistically similar. Otherwise, they are divided into different windows, as shown in the Time-Step Merging Unit of Figure~\ref{fig:structure}. The complete \(\mathcal{M}\) process is detailed in Algorithm~\ref{Algorithm:Adaptive Error Set Merging}.

\begin{algorithm}
\caption{Dual-Splitting Conformal Prediction (DSCP)}
\label{Algorithm:DSCP}
\begin{algorithmic}[1]
\setstretch{1.1}
\State \textbf{Input:} A trained point model \(f\) producing \(b\)-step forecasts, calibration dataset \(\{Z_i, K_i\}_{i=1}^m \), maximum number of categories \(N\), target error rate \(\alpha\), similarity threshold \( \Theta \), test set input \(Z_\tau=x_{\tau-a:\tau}\)
\State \textbf{Output:} \({{\hat{C}^{\alpha}_\tau}}\) for \(Z_\tau\)

\State \% Cluster Predictions
\State \(\{\hat{K}_i\}^m_{i=1} \leftarrow f(\{Z_i\}^m_{i=1})\)
\State \(G_{\hat{K}_1}, \ldots, G_{\hat{K}_m} \leftarrow \psi(\{\hat{K}_i\}^m_{i=1},~N)\)
\State \% Obtain \(\xi\)
\For{$i=1$ to $m$}
    \For{$j = 1$ to $b$}
        \State  \(\xi_{i,j} = [K_i-\hat{K}_i][j-1]\)
    \EndFor
\EndFor

\State \% Build Initial Error Set
\For{\(i = 1\) to \(m\)}
    \For{$j=1$ to $b$}
        \State \(\mathcal{E}_{i}' \leftarrow \bigcup_{t \in T_{K_i}} \{\xi_{t,j}\}\)
    \EndFor
\EndFor

\State \% Merge Similar Error Sets for Each Cluster
\For{\(i = 1\) to \(m\)}
    \State \({\tilde{\mathcal{E}}_{i,(1)}}, \ldots, {\tilde{\mathcal{E}}_{i,(b)}} \leftarrow \mathcal{M}(\Theta, \{\mathcal{E}'_{i,(1)}, \ldots, \mathcal{E}'_{i,(b)}\})\)
\EndFor

\State \% Get Confidence Interval \(\hat{{C^{\alpha}}}\)
\State For the new input \(Z_\tau\)
\State \(\hat{K_\tau} \leftarrow f(Z_\tau)=\hat{y}_{\tau+1:\tau+b}\)
\State \(R_\tau = \operatorname{argsort}\left(\operatorname{soft\_dtw}(\hat{K_\tau}, \{K_i\}_{i=1}^m)\right)\)
\State \(G_{\hat{K_\tau}} = \operatorname{majority} \left( G_{K_l} \mid l \in R_\tau[1:s] \right)\)
\For{$i=1$ to $m$}
    \If{\(G_{\hat{K_\tau}}=G_{K_i}\)}
        \State \(\beta \leftarrow i\)
    \EndIf
\EndFor
\State \({{\hat{C}^{\alpha}_{\tau,upper}}}(Z_\tau)=\{\hat{y}_{\tau+1},\ldots, \hat{y}_{\tau+b}\} + Q_{1-\alpha/2}(\tilde{E}_{\beta})\)
\State \({{\hat{C}^{\alpha}_{\tau,lower}}}(Z_\tau)=\{\hat{y}_{\tau+1},\ldots, \hat{y}_{\tau+b}\} + Q_{\alpha/2}(\tilde{E}_{\beta})\)
\State \textbf{return} \({{\hat{C}^{\alpha}_{\tau}}}(Z_\tau)\)
\end{algorithmic}
\end{algorithm}

The Dual-Splitting Conformal Prediction algorithm, as outlined in Algorithm~\ref{Algorithm:DSCP}, begins by clustering historical predictions using k-means to group similar trends, corresponding to the clustering step where \(\{\hat{K}_i\}_{i=1}^m\) are assigned to clusters \(G_{\hat{K}_1}, \ldots, G_{\hat{K}_m}\). Next, error terms \(\xi_{i,j}\) are extracted for each time step, aligning with the algorithm's computation of \(\xi_{i,j} = [K_i - \hat{K}_i][j-1]\). These errors are adaptively merged within clusters using the Kolmogorov-Smirnov test to group time steps with consistent \(\xi\) distributions, matching the \(\mathcal{M}\) process in the algorithm. Finally, for new predictions, confidence intervals are constructed using quantiles of the merged error sets from the assigned cluster, corresponding to the interval construction step where \(\hat{C}^{\alpha}_\tau\) is derived from \(\tilde{\mathcal{E}}_{\beta,(1)}, \ldots, \tilde{\mathcal{E}}_{\beta,(b)}\). This structured approach ensures robust uncertainty quantification across diverse time series scenarios.

\begin{table*}[htbp] % Starred version of table for two-column spanning
\centering
\caption{Comparison of $\Delta$Cov (\%), PI-Width, and WS* metrics across models at three confidence levels for various datasets.}
\label{tab:indicators_comparison}
\resizebox{\textwidth}{!}{
\begin{tabular}{llccccccccc}
\toprule
\raisebox{2pt}{Dataset} & \raisebox{2pt}{Methods} & \multicolumn{3}{c}{{95$\%$ confidence}} & \multicolumn{3}{c}{{90$\%$ confidence}} & \multicolumn{3}{c}{{85$\%$ confidence}} \\
\cmidrule(r){3-5} \cmidrule(lr){6-8} \cmidrule(l){9-11}
& & $\Delta$Cov & PI-Width & WS* & $\Delta$Cov & PI-Width & WS* & $\Delta$Cov & PI-Width & WS* \\
\midrule
\multirow{10}{*}{Solar Power}
& DSCP*       & -0.32 & 65.81 & \textbf{95.14} & 0.54 & 50.39 & \textbf{75.98} & 1.08 & 40.83 & \textbf{65.27} \\
& HopCPT      & 1.07 & 88.36 & 124.52 & 2.29 & 59.19 & 92.77 & 3.23 & 43.84 & 76.11 \\
& CP          & 0.86 & 106.92 & 155.52 & 1.76 & 70.62 & 116.44 & 2.58 & 50.50 & 94.66 \\
& EnbPI       & 0.31 & 100.61 & 154.92 & 0.81 & 65.14 & 115.84 & 1.15 & 45.60 & 94.09 \\
& ACI         & 0.41 & 94.17 & 138.43 & 0.75 & 62.98 & 106.06 & 1.01 & 46.21 & 87.93 \\
& CFRNN       & 1.00 & 101.31 & 141.18 & 1.92 & 68.93 & 107.78 & 2.78 & 51.46 & 89.19 \\
& Copula      & 0.28 & 95.03 & 140.89 & 0.75 & 64.16 & 107.62 & 1.17 & 47.33 & 88.99 \\
& CQR         & 0.61 & 122.82 & 130.76 & 4.19 & 92.38 & 97.79 & 7.48 & 82.11 & 86.48 \\
\midrule
\multirow{10}{*}{Spaflux}
& DSCP*       & -4.32 & 3134.99 & \textbf{5038.21} & 0.21 & 3041.41 & \textbf{4071.30} & 1.62 & 2584.69 & \textbf{3546.99} \\
& HopCPT      & -0.60 & 6425.99 & 7672.50 & 0.37 & 5291.87 & 6393.16 & 1.27 & 4446.24 & 5538.19 \\
& CP          & -1.80 & 4586.51 & 6193.88 & -2.02 & 3645.61 & 5171.40 & -2.37 & 3026.83 & 4573.66 \\
& EnbPI       & -1.14 & 4662.08 & 6167.98 & -1.26 & 3684.29 & 5149.67 & -1.47 & 3064.06 & 4557.65 \\
& ACI         & -0.27 & 4757.69 & 6062.33 & -0.38 & 3717.05 & 5116.65 & -0.37 & 3084.23 & 4540.87 \\
& CFRNN       & -3.59 & 4224.16 & 5925.54 & -3.95 & 3448.70 & 4959.42 & -4.09 & 2937.18 & 4398.94 \\
& Copula      & -1.95 & 4310.13 & 5653.39 & -2.20 & 3508.43 & 4814.13 & -2.34 & 2986.85 & 4307.84 \\
& CQR         & -2.04 & 6377.04 & 8371.76 & -7.11 & 3512.01 & 6332.31 & -7.44 & 3038.27 & 5371.05 \\
\midrule
\multirow{10}{*}{Air10}
& DSCP*       & -0.48 & 156.98 & \textbf{274.19} & -0.57 & 115.20 & \textbf{203.75} & -0.68 & 94.42 & \textbf{171.18} \\
& HopCPT      & -1.42 & 157.01 & 299.63 & -1.88 & 111.77 & 220.05 & -2.21 & 89.77 & 182.60 \\
& CP          & -2.08 & 147.37 & 300.56 & -3.29 & 105.45 & 221.86 & -4.07 & 84.20 & 183.99 \\
& EnbPI       & -1.36 & 156.89 & 301.23 & -1.94 & 111.81 & 221.85 & -2.30 & 89.54 & 183.91 \\
& ACI         & -0.55 & 168.71 & 293.08 & -0.87 & 118.93 & 218.14 & -1.08 & 94.39 & 181.90 \\
& CFRNN       & -2.57 & 143.57 & 302.18 & -3.75 & 103.74 & 221.79 & -4.42 & 83.37 & 183.59 \\
& Copula      & -1.50 & 154.85 & 299.40 & -2.04 & 111.09 & 220.47 & -2.32 & 89.31 & 182.76 \\
& CQR         & -0.48 & 227.18 & 337.85 & -0.57 & 151.75 & 241.21 & -1.34 & 112.64 & 192.65 \\
\midrule
\multirow{10}{*}{Wind Power}
& DSCP*       & -0.06 & 30.14 & \textbf{43.06} & 0.08 & 22.52 & \textbf{33.89} & 0.10 & 18.28 & \textbf{29.07} \\
& HopCPT      & -0.34 & 30.56 & 45.69 & -0.34 & 22.34 & 35.70 & -0.34 & 17.72 & 30.66 \\
& CP          & -0.68 & 30.24 & 47.05 & -1.07 & 21.78 & 36.95 & -1.39 & 17.15 & 31.37 \\
& EnbPI       & -0.48 & 30.58 & 46.82 & -0.68 & 22.10 & 36.78 & -0.85 & 17.44 & 31.25 \\
& ACI         & -0.03 & 31.80 & 46.78 & -0.04 & 22.98 & 36.81 & -0.04 & 18.13 & 31.30 \\
& CFRNN       & -0.69 & 29.90 & 46.31 & -1.10 & 21.66 & 36.49 & -1.44 & 17.09 & 31.06 \\
& Copula      & -0.49 & 30.25 & 46.01 & -0.70 & 22.01 & 36.29 & -0.88 & 17.40 & 30.91 \\
& CQR         & -6.01 & 36.68 & 58.83 & -4.46 & 26.27 & 39.94 & -2.64 & 23.59 & 34.36 \\
\bottomrule
\end{tabular}
}
\end{table*}

\section{Experiment}
In this section, comparative experiments are conducted on four types of time series datasets using different improved variants of CP for forecasting with multi-step CP methods. This section describes the datasets, models, benchmarks, and evaluation metrics used in the comparative analysis. The experimental results highlight DSCP's performance relative to other baseline methods.

\subsection{Setup}
\textbf{Datasets.}\ We use datasets from four different domains: (1) Solar radiation data from the National Solar Radiation Database (NSRDB) in the United States~\cite{sengupta2018national}. (2) An air quality dataset from Beijing~\cite{zhang2017cautionary} consisting of 12 time series from different monitoring stations over a 4-year period. This dataset includes two prediction targets, PM10 and PM2.5 concentrations, with a focus on PM10 data for our experiments. (3) Sap flow measurements from the Sapfluxnet data project~\cite{poyatos2021towards}. Due to high heterogeneity in the length of individual measurement series, we use a subset of 24 time series, each containing 15,000 to 20,000 data points with varying sampling rates. (4) Spatial Dynamic Wind Power Forecasting dataset provided by the Baidu KDD Cup 2022~\cite{baidu_aistudio_competition}. This dataset includes records collected from the Supervisory Control and Data Acquisition system of a wind farm, which consists of 134 wind turbines. The data are sampled every 10 minutes, covering a period of 245 days, with a total of 4,727,520 records. 

\textbf{Prediction Models.}\ We employ the Long Short-Term Memory (LSTM) model~\cite{hochreiter1997long} as the prediction model in the experiment. A global LSTM model was trained on all time series within the dataset, following the standards of state-of-the-art deep learning models (e.g., Oreshkin~\cite{oreshkin2019n}; Salinas~\cite{salinas2020deepar}; Smyl~\cite{smyl2020hybrid}). The global LSTM model was implemented with PyTorch~\cite{paszke2019pytorch}.

\textbf{Benchmarking Approaches.}\ We compare DSCP with different improved variants of CP, including: CP, HopCPT, EnbPI, ACI, CFRNN, CopulaCPTS, and CQR. Notably, among these methods, only the CQR method exclusively employs a QR predictive model for constructing confidence intervals, whereas the remaining baseline methods and DSCP rely on the same LSTM model. 

It is important to note that the compared methods were initially developed for uncertainty quantification in single-step forecasting. In this study, we focus on the problem of uncertainty quantification in multi-step forecasting. Therefore, to enable comparison with DSCP in a multi-step forecasting context, we adapt these methods to handle multi-step uncertainty quantification while preserving their core procedures. This adaptation allows us to evaluate the performance of DSCP against these adapted methods within the multi-step forecasting framework.

\textbf{Evaluation metrics.}\ To evaluate the performance of the methods, we employ three key metrics: $\Delta$Cov, PI-Width, and the Winkler Score (WS*). (1) $\Delta$Cov measures the deviation of the empirical coverage from the target confidence level, where smaller values indicate better performance. (2) PI-Width quantifies the average width of the prediction intervals, with smaller values being preferred. (3) WS* combines both coverage and interval width into a single measure, where lower values indicate higher accuracy and precision. The specific calculation of WS* is detailed in~\cite{auer2024conformal}.

\subsection{Improved Interval Accuracy: Experimental Analysis of DSCP's Classification Strategies}
\begin{figure*}[htbp]
    \centering
    \begin{minipage}{\textwidth}
        \centering
        \includegraphics[width=\textwidth]{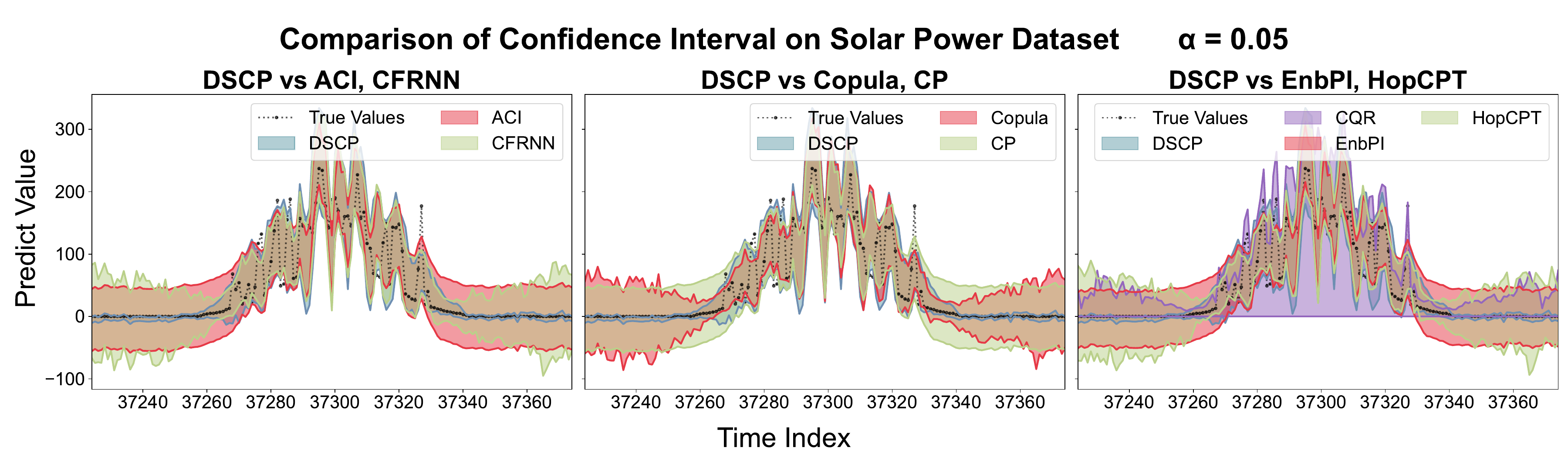}
        \label{fig:image1}
    \end{minipage}
    \begin{minipage}{\textwidth}
        \centering
        \includegraphics[width=\textwidth]{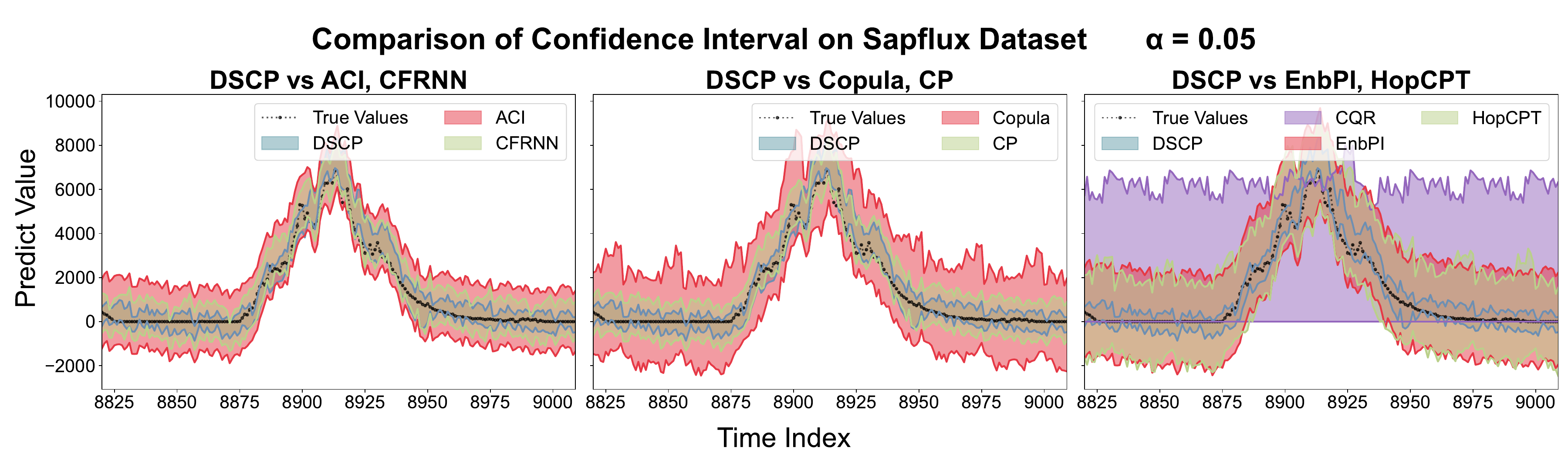}
        \label{fig:image2}
    \end{minipage}
    \begin{minipage}{\textwidth}
        \centering
        \includegraphics[width=\textwidth]{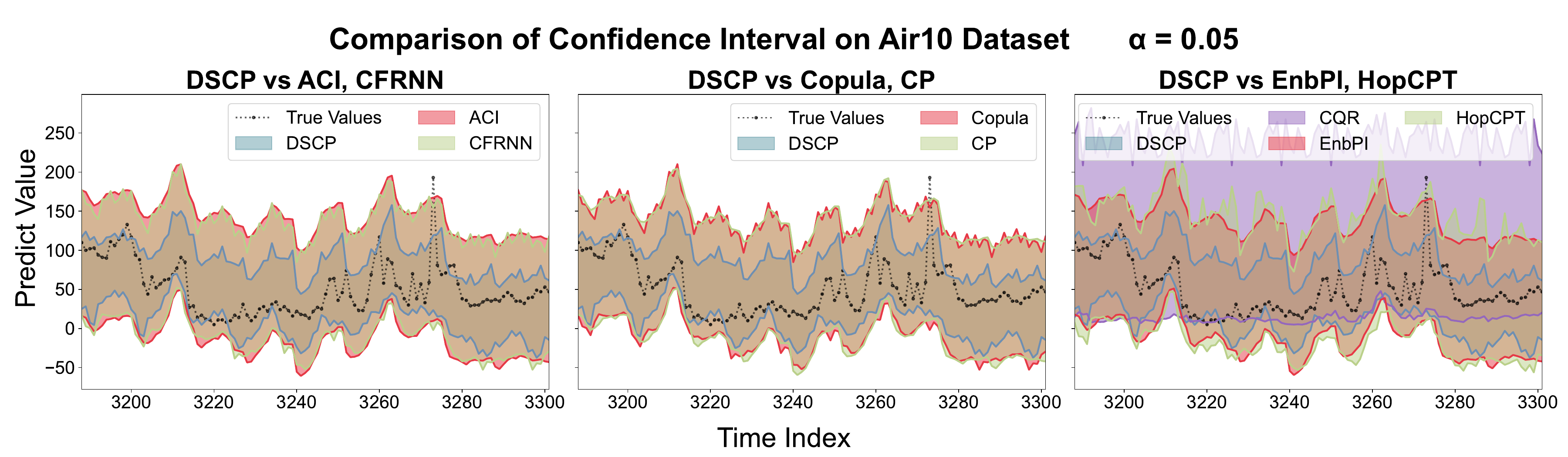}
        \label{fig:image3}
    \end{minipage}
    \begin{minipage}{\textwidth}
        \centering
        \includegraphics[width=\textwidth]{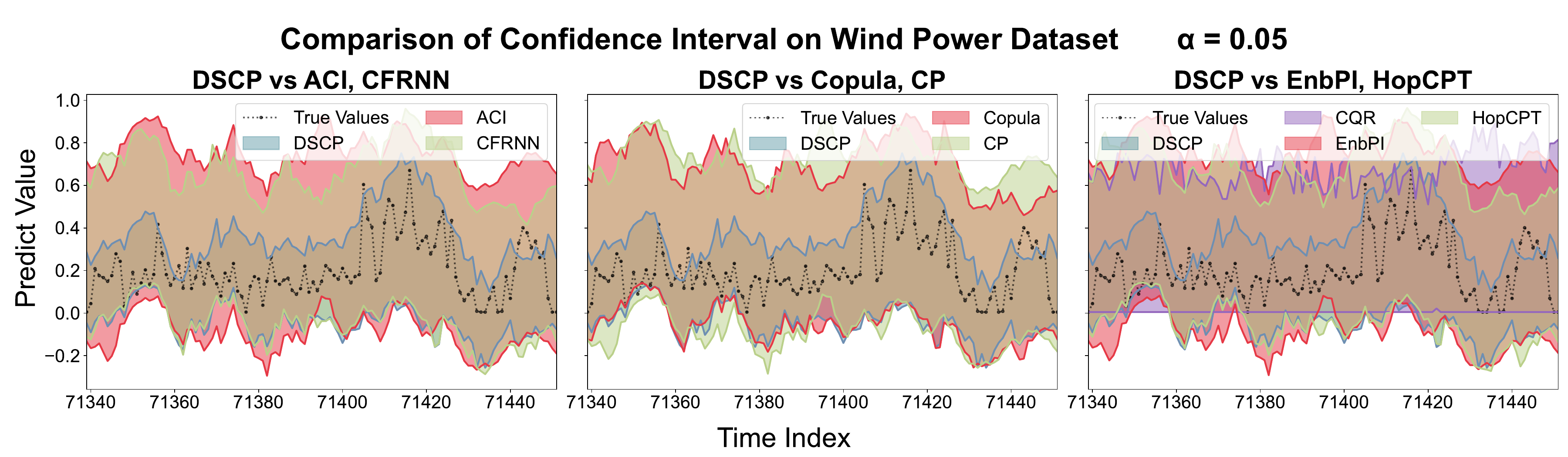}
        \label{fig:image4}
    \end{minipage}
    \caption{Comparison of confidence intervals generated by DSCP and other methods across four datasets at a confidence level of \(\alpha = 0.05\), demonstrating DSCP's ability to produce narrower and more accurate prediction intervals in varying data scenarios.}
    \label{fig:Experiment_result_figure}
\end{figure*}

\hyperref[tab:indicators_comparison]{Table~\ref*{tab:indicators_comparison}} compares the performance of various methods in constructing prediction intervals across multiple datasets at 95\%, 90\%, and 85\% confidence levels. DSCP achieves the best WS* across all datasets and confidence levels. The prediction intervals of the DSCP method, represented by the blue region in Fig.~\ref{fig:Experiment_result_figure}, are narrower and more accurate compared to other methods. This advantage enables DSCP to better capture the fluctuations of the true values, as demonstrated consistently across multiple datasets, including Solar Power, Sapflux, Air10, and Wind Power. \hyperref[tab:lianghua result]{Table~\ref*{tab:lianghua result}} further demonstrates DSCP's performance gains, with WS* improvements exceeding 20\% in some cases.

\begin{table}[htbp]
\centering
\caption{The following table presents the results of a quantitative analysis of the difference in Winkler scores between the DSCP and the most effective method in each dataset.}
\label{tab:lianghua result}
\begin{adjustbox}{width=1\textwidth}
\begin{tabular}{lccccccccc}
\toprule
& \multicolumn{3}{c}{95$\%$ confidence} & \multicolumn{3}{c}{90$\%$ confidence} & \multicolumn{3}{c}{85$\%$ confidence} \\
\cmidrule(r){2-4} \cmidrule(lr){5-7} \cmidrule(l){8-10}
{Dataset} & \makecell{DSCP\\WS*} & \makecell{Sota\\WS*} & {\makecell{Performance\\Gain}} & \makecell{DSCP\\WS*} & \makecell{Sota\\WS*} & {\makecell{Performance\\Gain}} & \makecell{DSCP\\WS*} & \makecell{Sota\\WS*} & {\makecell{Performance\\Gain}} \\
\midrule
\addlinespace[7pt]
Solar Power & 95.14 & 124.52 & \textcolor{red}{$\uparrow$ 23.59\%} & 75.98 & 92.77 & \textcolor{red}{$\uparrow$ 18.10\%} & 65.27 & 76.11 & \textcolor{red}{$\uparrow$ 14.24\%} \\
\addlinespace[5pt]
\midrule
\addlinespace[7pt]
Spaflux & 5038.21 & 5653.39 & \textcolor{red}{$\uparrow$ 10.88\%} & 4071.30 & 4814.13 & \textcolor{red}{$\uparrow$ 15.43\%} & 3546.99 & 4307.84 & \textcolor{red}{$\uparrow$ 17.66\%} \\
\addlinespace[5pt]
\midrule
\addlinespace[7pt]
Air10 & 275.62 & 293.08 & \textcolor{red}{$\uparrow$ 5.96\%} & 205.01 & 218.14 & \textcolor{red}{$\uparrow$ 6.02\%} & 172.24 & 181.90 & \textcolor{red}{$\uparrow$ 5.31\%} \\
\addlinespace[5pt]
\midrule
\addlinespace[7pt]
Wind Power & 43.06 & 45.69 & \textcolor{red}{$\uparrow$ 5.76\%} & 33.89 & 35.70 & \textcolor{red}{$\uparrow$ 5.07\%} & 29.07 & 30.66 & \textcolor{red}{$\uparrow$ 5.19\%} \\
\addlinespace[5pt]
\bottomrule
\end{tabular}
\end{adjustbox}
\end{table}

\textbf{Asymmetric Error Handling:} By obtaining \(\xi\) from~\eqref{eq:None_Absolute value solving epsilon}, DSCP effectively distinguishes between \(\xi_{over}\) and \(\xi_{under}\) errors, capturing the error distribution characteristics of point model \(f\) more accurately. This separation allows DSCP to generate asymmetric prediction intervals \(\hat{C}_i\) that better reflect the true data distribution. For example, in the Wind Power dataset, DSCP generates \(\hat{C}_i\) with noticeably narrower upper bounds \(\hat{C}_i^{(U)}\) compared to other methods. Specifically, the distance from \(\hat{Z}_i\) to \(\hat{C}_i^{(U)}\) is smaller for DSCP, reflecting the point predictor's tendency to overestimate true values. This asymmetry in \(\hat{C}_i\) improves both the sharpness and reliability of the intervals. In contrast, most other methods produce symmetric \(\hat{C}_i\) with equal widths, often resulting in overly wide bounds that fail to capture specific error characteristics, such as overestimation or underestimation tendencies.

\textbf{Magnitude and Trend Classification:} DSCP classifies \(\hat{K}_i\) based on magnitude and trend, preventing the mixing of \(\xi\) from different types of \(\hat{K}_i\). This classification ensures that error terms are applied only to predictions with similar characteristics, improving the accuracy of the constructed intervals. For example, in the Solar Power dataset, where prediction magnitudes vary significantly, DSCP ensures that larger \(\xi_{under}\) from high-magnitude \(\hat{K}_i\) are not incorrectly applied to low-magnitude \(\hat{K}_i\). Without this classification, \(\xi\) from high-magnitude predictions could lead to overly wide intervals for low-magnitude \(\hat{K}_i\), while \(\xi\) from low-magnitude predictions could unnecessarily narrow intervals for high-magnitude \(\hat{K}_i\). By applying errors only to appropriate \(\hat{K}_i\), DSCP improves interval \(\hat{C}\) accuracy and avoids misalignment between error terms and prediction characteristics.

\textbf{Dynamic Error Subset Merging:} DSCP dynamically merges error subsets \( \mathcal{E}'_{(1)}, \ldots, \mathcal{E}'_{(b)} \) based on time steps, ensuring that error subsets with similar uncertainty distributions are grouped together. This merging process is particularly beneficial for datasets with periodic characteristics, where uncertainty varies over time. For instance, in the Solar Power dataset, uncertainty differs significantly between day and night. Without dynamic merging, error term \(\xi_i\) from different time steps could be mixed, leading to inaccurate \(\hat{C}_i\). For example, \(\xi_i\) during the day, where uncertainty is higher, might be used to construct \(\hat{C}_i\) for nighttime predictions, resulting in overly wide intervals for night and vice versa. By dynamically merging error subsets, DSCP ensures that each time step's merged error subset \(\tilde{\mathcal{E}}_{(i)}\) contains only similar uncertainty distributions, improving the accuracy and reliability of \(\hat{C}_i\).

\subsection{Robustness to Data Periodicity: Sensitivity Analysis on Prediction Time Steps}
\begin{figure}[htbp]
\centering
\includegraphics[width=0.7\linewidth]{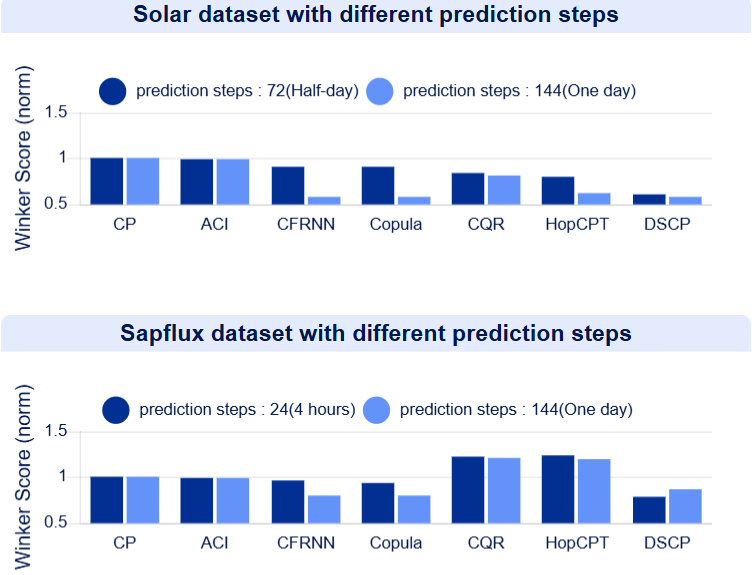}
\caption{Comparison of DSCP and other CP methods across different prediction steps, demonstrating DSCP's robustness to varying prediction steps without reliance on data periodicity.}
\label{fig:Prediction_Time_Steps}
\end{figure}

To further evaluate DSCP's robustness to data periodicity, we conducted a sensitivity analysis by varying the prediction time steps. Fig.~\ref{fig:Prediction_Time_Steps} compares the performance of DSCP and other CP methods across different prediction steps. DSCP maintains consistent performance regardless of data periodicity, with particularly strong results in the Solar dataset. In contrast, other methods show notable performance degradation when prediction steps do not align with the data periodicity. This is particularly important because the periodicity of time-series data is often difficult to determine or may be unstable in real-world applications. DSCP's ability to adapt to varying prediction steps without relying on precise periodicity information makes it a more reliable choice for multi-step forecasting tasks.

In summary, these multi-steps experimental results show that the DSCP method is capable of constructing prediction intervals more accurately through error sign differentiation, classification of prediction results, and dynamic merging on time steps. This not only makes DSCP more flexible in dealing with different uncertainty distributions, but also improves its ability to adapt to cyclical changes in data, thus showing significant advantages in multi-step time series forecasting tasks.

\section{Case Study}
\begin{figure}[htbp]
\centering
\includegraphics[width=0.8\linewidth]{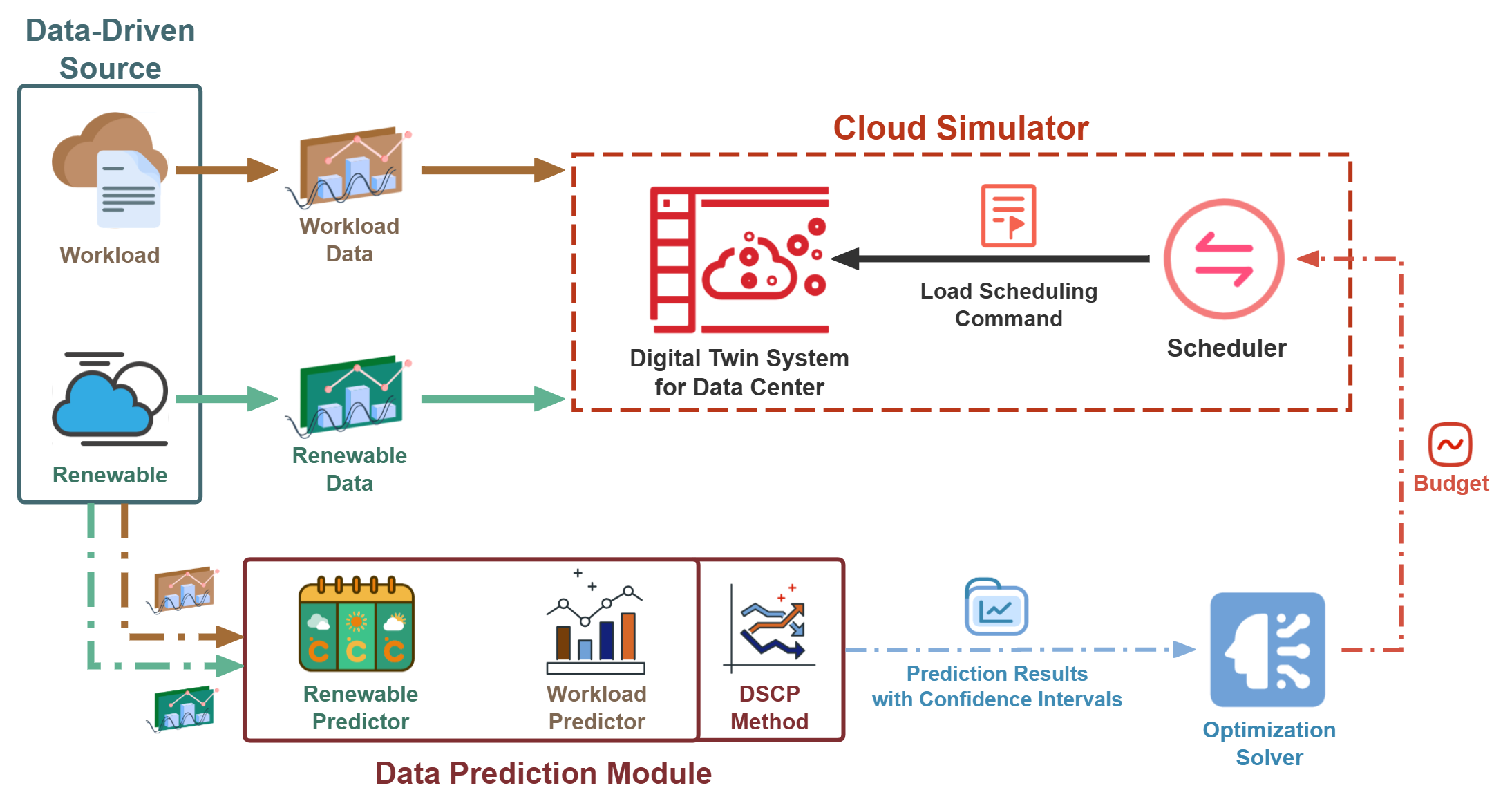}
\caption{Workflow of the energy optimization case study, showing data flow from sources to the prediction module, optimization solver, and cloud simulator. Multi-step prediction intervals generated by DSCP guide task scheduling, which is evaluated for energy consumption and performance.}
\label{fig:Case study}
\end{figure}

The case study simulates energy optimization in a data center using predictive control methods. As illustrated in Fig.~\ref{fig:Case study}, the workflow integrates four main components: data sources, a prediction module, an optimization solver, and a cloud simulator. Data sources provide real-time workload and renewable energy data, which are processed by the prediction module to generate multi-step prediction intervals using DSCP. The optimization solver uses these predictions to compute scheduling strategies, while the cloud simulator evaluates the system's energy consumption and performance under these strategies. This simulation approach demonstrates the energy optimization capability of the current strategy and therefore enables a quantitative comparison of the performance improvements achieved by the DSCP.

\subsection{Optimization Problem \& Method}
The core optimization problem is to minimize carbon emissions by aligning task scheduling with renewable energy availability within a predicted time window of size \( n \). At the beginning of the \( \tau \)-th time step, the optimization is formulated as:
\begin{equation}
\underset{W_\tau^D, \ldots, W_{\tau+n}^D, W_{dely}^D}{\text{minimize}} \sum_{k=\tau}^{\tau+n} \varphi_k \cdot P_k \cdot \gamma^{k-\tau} + D \cdot \gamma^n,
\end{equation}
where the subscript \( k \) denotes the value at time step \( k \), \( \varphi_k = \max \left( W_k^D + W_k^{L} - W_k^{*Re}, 0 \right) \) represents the additional brown energy (non-renewable energy, e.g., fossil fuels) required to meet the load demand, with \( W_k^D \) as the load allocated, \( W_k^{L} \) the load already running, \( W_k^{*Re} \) the predicted available renewable energy, \( P_k \) the brown energy price, and \( \gamma^{k-\tau} \) a time decay factor assigning higher weights to near-term costs due to lower uncertainty. The term \( D = W_{dely}^D \cdot P_{max} \) captures the cost of delaying tasks beyond the planning window, where \( W_{dely}^D \) represents the load delayed to the next optimization window, and \( P_{max} \) denotes the highest brown energy price, penalizing excessive delays. The optimization is subject to the constraint:
\begin{equation}
\sum_{k=\tau}^{\tau+n} W_k^D + W_{dely}^D = \sum_{k=\tau}^{\tau+n} W_k^{*Load} + W_{\tau}^{Dely},
\end{equation}
which ensures that the total allocated load matches the total predicted workload \( W_k^{*Load} \) and the load \( W_{\tau}^{Dely} \) carried over from the previous optimization window, maintaining system balance and avoiding resource overcommitment.

By incorporating confidence intervals, scheduling strategies become more flexible, enabling dynamic balancing of efficiency and robustness based on predicted uncertainty levels. To achieve this, DSCP replaces the point predictions \( W_k^{*Re} \) and \( W_k^{*Load} \) in the optimization problem with multi-step prediction intervals for workload and renewable energy, providing richer information than single-step predictions. These intervals include the range of possible outcomes and their probabilities at each time step, enabling the solver to evaluate risks and trade-offs more effectively. For instance, wider intervals indicating higher uncertainty prompt conservative strategies, such as allocating backup resources to avoid excessive task backlog. Conversely, narrower intervals indicating lower uncertainty support aggressive optimization, such as maximizing renewable energy use while reasonably deferring tasks, thereby reducing carbon emissions.

\begin{figure}[htbp]
    \centering
    \includegraphics[width=0.7\linewidth]{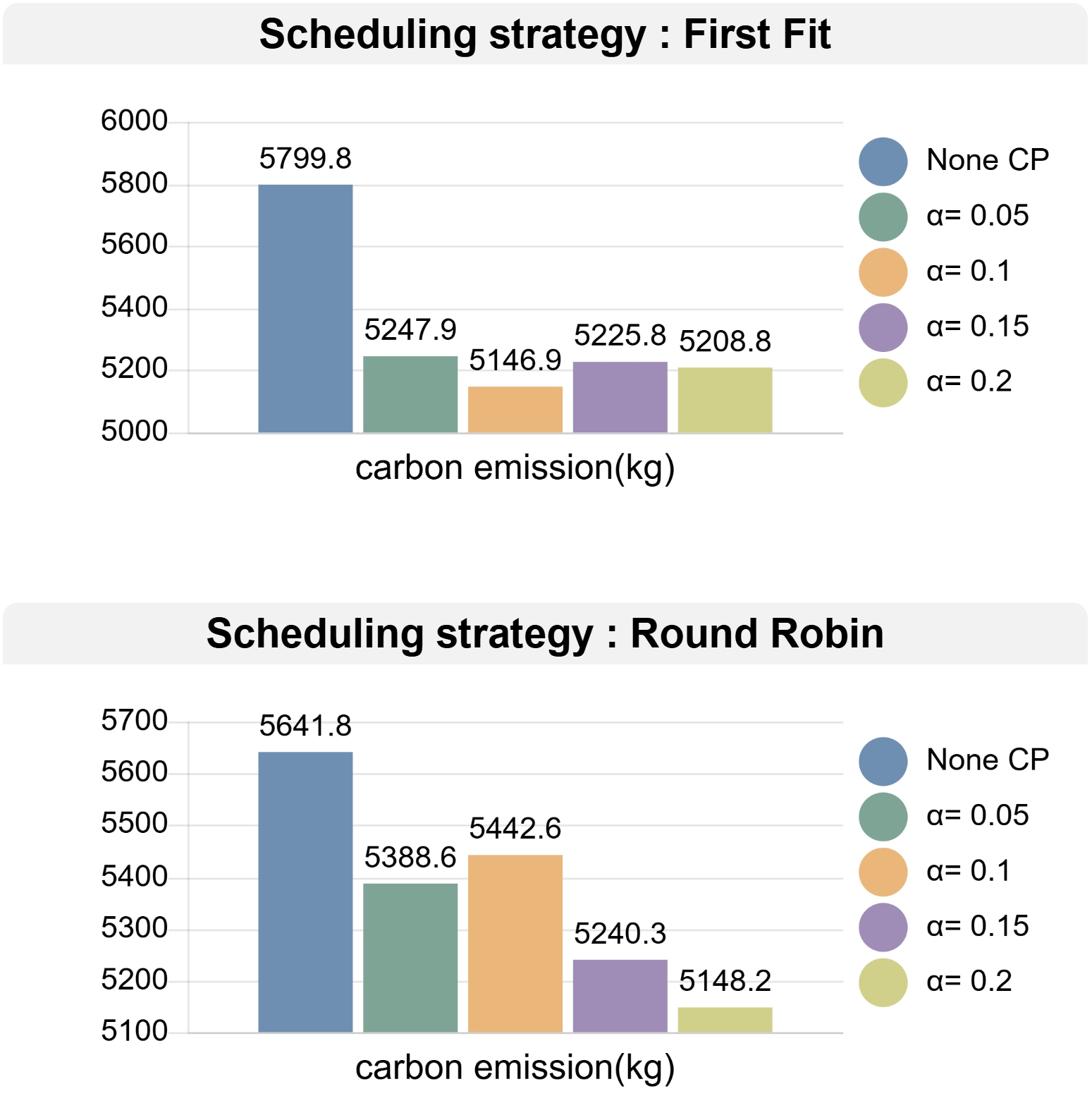}
    \caption{Comparison of carbon emissions under First Fit and Round Robin scheduling strategies with DSCP at different confidence levels (\(\alpha = 0.05, 0.1, 0.15, 0.2\)). The results demonstrate DSCP's effectiveness in reducing emissions and improving renewable energy utilization, with RR showing a stronger dependence on confidence level compared to FF.}
    \label{fig:Case Study result}
\end{figure}

\subsection{Results}
We tested DSCP under different confidence levels (\(\alpha = 0.05, 0.1, 0.15, 0.2\)) and compared its performance with a baseline scenario without conformal prediction (None CP). The results, shown in Fig.~\ref{fig:Case Study result}, illustrate DSCP's impact on carbon emissions(kg) under two scheduling strategies: First Fit (FF) and Round Robin (RR). DSCP consistently reduces carbon emissions across all confidence levels, achieving an average reduction of approximately 8.05\% relative to the baseline. The most significant improvement is observed under the FF strategy, where emissions decrease by up to 11.25\% at \(\alpha = 0.1\). These results highlight DSCP's effectiveness in enhancing energy efficiency and reducing environmental impact.

In conclusion, DSCP’s integration of uncertainty information offers a robust approach to adaptive decision-making in complex systems. This capability demonstrates its potential for enabling efficient resource allocation and supporting sustainable operational strategies.

\section{Conclusion and Future Work}
In conclusion, we propose the DSCP, which utilizes vertical clustering and horizontal merging to construct error sets for accurate uncertainty quantification in multi-step time series forecasting. Compared to existing baselines, DSCP achieves a maximum performance improvement of 23.59\%. Moreover, we integrate DSCP into data center energy management case, helping the original First Fit scheduling strategy reduce carbon emissions by up to 11.25\%, demonstrating its practical value in real-world applications.

Future work will focus on three key directions. First, we aim to enhance the vertical clustering and horizontal merging processes by incorporating more advanced machine learning models, which can capture finer-grained temporal patterns and further improve DSCP's performance. Second, we will explore whether uncertainty information can be transferred across related prediction tasks, potentially enabling DSCP to leverage knowledge from one domain to improve performance in another. Finally, we plan to optimize DSCP for large-scale datasets and distributed computing environments, significantly enhancing its scalability and making it suitable for broader real-world applications. These efforts will further bridge the gap between theoretical innovation and practical impact, advancing the field of uncertainty-aware forecasting.

% \printbibliography
\bibliographystyle{unsrt}
\bibliography{Applied_Soft_Computing}

\begin{thebibliography}{10}

\bibitem{liu2024short}
Xin Liu and Jun Zhou.
\newblock Short-term wind power forecasting based on multivariate/multi-step
  lstm with temporal feature attention mechanism.
\newblock {\em Applied Soft Computing}, 150:111050, 2024.

\bibitem{dolgintseva2024comparison}
E~Dolgintseva, H~Wu, O~Petrosian, A~Zhadan, A~Allakhverdyan, and A~Martemyanov.
\newblock Comparison of multi-step forecasting methods for renewable energy.
\newblock {\em Energy Systems}, pages 1--32, 2024.

\bibitem{ghobadi2022multi}
Fatemeh Ghobadi and Doosun Kang.
\newblock Multi-step ahead probabilistic forecasting of daily streamflow using
  bayesian deep learning: A multiple case study.
\newblock {\em Water}, 14(22):3672, 2022.

\bibitem{yang2021comprehensive}
Siyue Yang and Yukun Bao.
\newblock Comprehensive learning particle swarm optimization enabled modeling
  framework for multi-step-ahead influenza prediction.
\newblock {\em Applied Soft Computing}, 113:107994, 2021.

\bibitem{priya2019resource}
V~Priya, C~Sathiya Kumar, and Ramani Kannan.
\newblock Resource scheduling algorithm with load balancing for cloud service
  provisioning.
\newblock {\em Applied Soft Computing}, 76:416--424, 2019.

\bibitem{he2024data}
Renfei He, Limao Zhang, and Alvin Wei~Ze Chew.
\newblock Data-driven multi-step prediction and analysis of monthly rainfall
  using explainable deep learning.
\newblock {\em Expert Systems with Applications}, 235:121160, 2024.

\bibitem{zhang2021multi}
Ning Zhang and Alice Alipour.
\newblock A multi-step assessment framework for optimization of flood
  mitigation strategies in transportation networks.
\newblock {\em International Journal of Disaster Risk Reduction}, 63:102439,
  2021.

\bibitem{lim2021time}
Bryan Lim and Stefan Zohren.
\newblock Time-series forecasting with deep learning: a survey.
\newblock {\em Philosophical Transactions of the Royal Society A},
  379(2194):20200209, 2021.

\bibitem{zhou2018predicting}
Xian Zhou, Yanyan Shen, Yanmin Zhu, and Linpeng Huang.
\newblock Predicting multi-step citywide passenger demands using
  attention-based neural networks.
\newblock In {\em Proceedings of the Eleventh ACM international conference on
  web search and data mining}, pages 736--744, 2018.

\bibitem{abdar2021review}
Moloud Abdar, Farhad Pourpanah, Sadiq Hussain, Dana Rezazadegan, Li~Liu,
  Mohammad Ghavamzadeh, Paul Fieguth, Xiaochun Cao, Abbas Khosravi, U~Rajendra
  Acharya, et~al.
\newblock A review of uncertainty quantification in deep learning: Techniques,
  applications and challenges.
\newblock {\em Information fusion}, 76:243--297, 2021.

\bibitem{kabir2018neural}
HM~Dipu Kabir, Abbas Khosravi, Mohammad~Anwar Hosen, and Saeid Nahavandi.
\newblock Neural network-based uncertainty quantification: A survey of
  methodologies and applications.
\newblock {\em IEEE access}, 6:36218--36234, 2018.

\bibitem{almeida2015prediction}
V{\^a}nia Almeida and Joao Gama.
\newblock Prediction intervals for electric load forecast: Evaluation for
  different profiles.
\newblock In {\em 2015 18th International Conference on Intelligent System
  Application to Power Systems (ISAP)}, pages 1--6. IEEE, 2015.

\bibitem{quan2014uncertainty}
Hao Quan, Dipti Srinivasan, and Abbas Khosravi.
\newblock Uncertainty handling using neural network-based prediction intervals
  for electrical load forecasting.
\newblock {\em Energy}, 73:916--925, 2014.

\bibitem{quan2013short}
Hao Quan, Dipti Srinivasan, and Abbas Khosravi.
\newblock Short-term load and wind power forecasting using neural network-based
  prediction intervals.
\newblock {\em IEEE transactions on neural networks and learning systems},
  25(2):303--315, 2013.

\bibitem{tseng2010comparing}
Fang-Mei Tseng and Yi-Chung Hu.
\newblock Comparing four bankruptcy prediction models: Logit, quadratic
  interval logit, neural and fuzzy neural networks.
\newblock {\em Expert systems with applications}, 37(3):1846--1853, 2010.

\bibitem{zhang2007statistical}
Yan-Qing Zhang and Xuhui Wan.
\newblock Statistical fuzzy interval neural networks for currency exchange rate
  time series prediction.
\newblock {\em Applied Soft Computing}, 7(4):1149--1156, 2007.

\bibitem{pinson2010conditional}
Pierre Pinson and George Kariniotakis.
\newblock Conditional prediction intervals of wind power generation.
\newblock {\em IEEE Transactions on Power Systems}, 25(4):1845--1856, 2010.

\bibitem{galvan2017multi}
In{\'e}s~M Galv{\'a}n, Jos{\'e}~M Valls, Alejandro Cervantes, and Ricardo Aler.
\newblock Multi-objective evolutionary optimization of prediction intervals for
  solar energy forecasting with neural networks.
\newblock {\em Information Sciences}, 418:363--382, 2017.

\bibitem{wang2023conformal}
Wei Wang, Bin Feng, Gang Huang, Chuangxin Guo, Wenlong Liao, and Zhe Chen.
\newblock Conformal asymmetric multi-quantile generative transformer for
  day-ahead wind power interval prediction.
\newblock {\em Applied Energy}, 333:120634, 2023.

\bibitem{nishiura2012early}
Hiroshi Nishiura.
\newblock Early detection of nosocomial outbreaks caused by rare pathogens: a
  case study employing score prediction interval.
\newblock {\em Osong public health and research perspectives}, 3(3):121--127,
  2012.

\bibitem{zee2016stroke}
Benny Zee, Jack Lee, Qing Li, Vincent Mok, Alice Kong, Lap-Kin Chiang, Lorna
  Ng, Yuanyuan Zhuo, H~Yu, and Z~Yang.
\newblock Stroke risk assessment for the community by automatic retinal image
  analysis using fundus photograph.
\newblock {\em Qual Primary Care}, 24(3):114--124, 2016.

\bibitem{hernandez2020uncertainty}
S~Hern{\'a}ndez and Juan~L L{\'o}pez.
\newblock Uncertainty quantification for plant disease detection using bayesian
  deep learning.
\newblock {\em Applied Soft Computing}, 96:106597, 2020.

\bibitem{shafer2008tutorial}
Glenn Shafer and Vladimir Vovk.
\newblock A tutorial on conformal prediction.
\newblock {\em Journal of Machine Learning Research}, 9(3), 2008.

\bibitem{fontana2023conformal}
Matteo Fontana, Gianluca Zeni, and Simone Vantini.
\newblock Conformal prediction: a unified review of theory and new challenges.
\newblock {\em Bernoulli}, 29(1):1--23, 2023.

\bibitem{box2011bayesian}
George~EP Box and George~C Tiao.
\newblock {\em Bayesian inference in statistical analysis}.
\newblock John Wiley \& Sons, 2011.

\bibitem{xu2021conformal}
Chen Xu and Yao Xie.
\newblock Conformal prediction interval for dynamic time-series.
\newblock In {\em International Conference on Machine Learning}, pages
  11559--11569. PMLR, 2021.

\bibitem{barber2023conformal}
Rina~Foygel Barber, Emmanuel~J Candes, Aaditya Ramdas, and Ryan~J Tibshirani.
\newblock Conformal prediction beyond exchangeability.
\newblock {\em The Annals of Statistics}, 51(2):816--845, 2023.

\bibitem{gibbs2021adaptive}
Isaac Gibbs and Emmanuel Candes.
\newblock Adaptive conformal inference under distribution shift.
\newblock {\em Advances in Neural Information Processing Systems},
  34:1660--1672, 2021.

\bibitem{sesia2021conformal}
Matteo Sesia and Yaniv Romano.
\newblock Conformal prediction using conditional histograms.
\newblock {\em Advances in Neural Information Processing Systems},
  34:6304--6315, 2021.

\bibitem{bastani2022practical}
Osbert Bastani, Varun Gupta, Christopher Jung, Georgy Noarov, Ramya Ramalingam,
  and Aaron Roth.
\newblock Practical adversarial multivalid conformal prediction.
\newblock {\em Advances in Neural Information Processing Systems},
  35:29362--29373, 2022.

\bibitem{lin2022conformal}
Zhen Lin, Shubhendu Trivedi, and Jimeng Sun.
\newblock Conformal prediction with temporal quantile adjustments.
\newblock {\em Advances in Neural Information Processing Systems},
  35:31017--31030, 2022.

\bibitem{angelopoulos2024conformal}
Anastasios Angelopoulos, Emmanuel Candes, and Ryan~J Tibshirani.
\newblock Conformal pid control for time series prediction.
\newblock {\em Advances in neural information processing systems}, 36, 2024.

\bibitem{koenker1978regression}
Roger Koenker and Gilbert Bassett~Jr.
\newblock Regression quantiles.
\newblock {\em Econometrica: journal of the Econometric Society}, pages 33--50,
  1978.

\bibitem{xu2023sequential}
Chen Xu and Yao Xie.
\newblock Sequential predictive conformal inference for time series.
\newblock In {\em International Conference on Machine Learning}, pages
  38707--38727. PMLR, 2023.

\bibitem{stankeviciute2021conformal}
Kamile Stankeviciute, Ahmed M~Alaa, and Mihaela van~der Schaar.
\newblock Conformal time-series forecasting.
\newblock {\em Advances in neural information processing systems},
  34:6216--6228, 2021.

\bibitem{sun2022copula}
Sophia Sun and Rose Yu.
\newblock Copula conformal prediction for multi-step time series forecasting.
\newblock {\em arXiv preprint arXiv:2212.03281}, 2022.

\bibitem{auer2024conformal}
Andreas Auer, Martin Gauch, Daniel Klotz, and Sepp Hochreiter.
\newblock Conformal prediction for time series with modern hopfield networks.
\newblock {\em Advances in Neural Information Processing Systems}, 36, 2024.

\bibitem{romano2019conformalized}
Yaniv Romano, Evan Patterson, and Emmanuel Candes.
\newblock Conformalized quantile regression.
\newblock {\em Advances in neural information processing systems}, 32, 2019.

\bibitem{peterson2009k}
Leif~E Peterson.
\newblock K-nearest neighbor.
\newblock {\em Scholarpedia}, 4(2):1883, 2009.

\bibitem{muller2007dynamic}
Meinard M{\"u}ller.
\newblock Dynamic time warping.
\newblock {\em Information retrieval for music and motion}, pages 69--84, 2007.

\bibitem{rabiner1986introduction}
Lawrence Rabiner and Biinghwang Juang.
\newblock An introduction to hidden markov models.
\newblock {\em ieee assp magazine}, 3(1):4--16, 1986.

\bibitem{yamashita2018convolutional}
Rikiya Yamashita, Mizuho Nishio, Richard Kinh~Gian Do, and Kaori Togashi.
\newblock Convolutional neural networks: an overview and application in
  radiology.
\newblock {\em Insights into imaging}, 9:611--629, 2018.

\bibitem{graves2012long}
Alex Graves and Alex Graves.
\newblock Long short-term memory.
\newblock {\em Supervised sequence labelling with recurrent neural networks},
  pages 37--45, 2012.

\bibitem{ahmed2020k}
Mohiuddin Ahmed, Raihan Seraj, and Syed Mohammed~Shamsul Islam.
\newblock The k-means algorithm: A comprehensive survey and performance
  evaluation.
\newblock {\em Electronics}, 9(8):1295, 2020.

\bibitem{berger2014kolmogorov}
Vance~W Berger and YanYan Zhou.
\newblock Kolmogorov--smirnov test: Overview.
\newblock {\em Wiley statsref: Statistics reference online}, 2014.

\bibitem{macqueen1967some}
James MacQueen et~al.
\newblock Some methods for classification and analysis of multivariate
  observations.
\newblock In {\em Proceedings of the fifth Berkeley symposium on mathematical
  statistics and probability}, volume~1, pages 281--297. Oakland, CA, USA,
  1967.

\bibitem{rousseeuw1987silhouettes}
Peter~J Rousseeuw.
\newblock Silhouettes: a graphical aid to the interpretation and validation of
  cluster analysis.
\newblock {\em Journal of computational and applied mathematics}, 20:53--65,
  1987.

\bibitem{cuturi2017soft}
Marco Cuturi and Mathieu Blondel.
\newblock Soft-dtw: a differentiable loss function for time-series.
\newblock In {\em International conference on machine learning}, pages
  894--903. PMLR, 2017.

\bibitem{sengupta2018national}
Manajit Sengupta, Yu~Xie, Anthony Lopez, Aron Habte, Galen Maclaurin, and James
  Shelby.
\newblock The national solar radiation data base (nsrdb).
\newblock {\em Renewable and sustainable energy reviews}, 89:51--60, 2018.

\bibitem{zhang2017cautionary}
Shuyi Zhang, Bin Guo, Anlan Dong, Jing He, Ziping Xu, and Song~Xi Chen.
\newblock Cautionary tales on air-quality improvement in beijing.
\newblock {\em Proceedings of the Royal Society A: Mathematical, Physical and
  Engineering Sciences}, 473(2205):20170457, 2017.

\bibitem{poyatos2021towards}
Rafael Poyatos, V{\'\i}ctor Granda, V{\'\i}ctor Flo, Maurizio Mencuccini, and
  Jordi Mart{\'\i}nez-Vilalta.
\newblock Towards a consistent quantification of ecosystem transpiration and
  its uncertainty from the sapfluxnet database.
\newblock In {\em EGU General Assembly Conference Abstracts}, pages
  EGU21--13326, 2021.

\bibitem{baidu_aistudio_competition}
Baidu~AI Studio.
\newblock Data competition: Paddlepaddle, national weather forecasting.
\newblock
  \url{https://aistudio.baidu.com/competition/detail/152/0/introduction}.
\newblock Accessed: 2024-11-15.

\bibitem{hochreiter1997long}
Sepp Hochreiter and J{\"u}rgen Schmidhuber.
\newblock Long short-term memory.
\newblock {\em Neural computation}, 9(8):1735--1780, 1997.

\bibitem{oreshkin2019n}
Boris~N Oreshkin, Dmitri Carpov, Nicolas Chapados, and Yoshua Bengio.
\newblock N-beats: Neural basis expansion analysis for interpretable time
  series forecasting.
\newblock {\em arXiv preprint arXiv:1905.10437}, 2019.

\bibitem{salinas2020deepar}
David Salinas, Valentin Flunkert, Jan Gasthaus, and Tim Januschowski.
\newblock Deepar: Probabilistic forecasting with autoregressive recurrent
  networks.
\newblock {\em International journal of forecasting}, 36(3):1181--1191, 2020.

\bibitem{smyl2020hybrid}
Slawek Smyl.
\newblock A hybrid method of exponential smoothing and recurrent neural
  networks for time series forecasting.
\newblock {\em International journal of forecasting}, 36(1):75--85, 2020.

\bibitem{paszke2019pytorch}
Adam Paszke, Sam Gross, Francisco Massa, Adam Lerer, James Bradbury, Gregory
  Chanan, Trevor Killeen, Zeming Lin, Natalia Gimelshein, Luca Antiga, et~al.
\newblock Pytorch: An imperative style, high-performance deep learning library.
\newblock {\em Advances in neural information processing systems}, 32, 2019.

\end{thebibliography}

\end{document}